*Article*

# Effects of Image Size on Deep Learning

**Olivier Rukundo**

Clinical Physiology (Lund), Department of Clinical Sciences (Lund), Lund University,
Skåne University Hospital (2nd Floor, Hall B), 221 85 Lund, Sweden: orukundo@gmail.com

**Abstract:** In this work, the best size for late gadolinium enhancement (LGE) magnetic resonance imaging (MRI) images in the training dataset was determined to optimize deep learning training outcomes. Non-extra pixel and extra pixel interpolation algorithms were used to determine the new size of the LGE-MRI images. A novel strategy was introduced to handle interpolation masks and remove extra class labels in interpolated ground truth (GT) segmentation masks. The expectation maximization, weighted intensity, a priori information (EWA) algorithm was used for quantification of myocardial infarction (MI) in automatically segmented LGE-MRI images. Arbitrary threshold, comparison of the sums, and sums of differences are methods used to estimate the relationship between semi-automatic or manual and fully automated quantification of myocardial infarction (MI) results. The relationship between semi-automatic and fully automated quantification of MI results was found to be closer in the case of bigger LGE MRI images (55.5% closer to manual results) than in the case of smaller LGE MRI images (22.2% closer to manual results).

## 1. Introduction

In this study, the main objective is to determine the best size of LGE-MRI images in the training datasets to achieve optimal deep learning-based segmentation outcomes. Deep learning is a subfield of machine learning and refers to a particular class of neural networks [1], [2], [3],[4],[5]. Neural networks are the backbone of deep learning algorithms and un-like shallow counterparts, deep neural networks can directly process raw input data, including images, text, and sound [5]. In deep learning, a class of deep neural networks commonly applied to visual imagery is CNN [3], [5], [6]. Figure 1 shows a simplified representation of a few common deep learning architectures, applicable to visual imagery [9]. Figure 1 shows a schematic representation of two examples of the most commonly used networks. As can be seen, in Figure 1, one type of deep neural network architecture can also form the backbone of more sophisticated architectures for advanced applications [5], [7], [8], [9]. In this paper, the CNN architecture of interest is U-net. U-net was chosen not only because it outperformed the then-best method of sliding-window convolutional network or won many challenges but also because it could provide a fast and precise segmentation of heart images [10]. Typically, image segmentation locates object boundaries in the image to simplify or change the image into something more meaningful and/or easier to analyse [11], [12], [13], [14], [15]. In medical image analysis, segmentation is the stage where a significant commitment is made to delineate structures of interest and discriminate them from background tissue, but this kind of separation or segmentation is generally effortless and swift for the human visual system [16], [17], [18], [19]. In this work, U-net was dedicated to that stage to ensure swift and accurate delineations and discriminations.

The current literature shows that there exist many works which are mostly proposed for segmentation of medical images using U-net or closely related versions [20], [21], [22], [23], [24], [25], [26], [44], [45], [48], [49], [50].

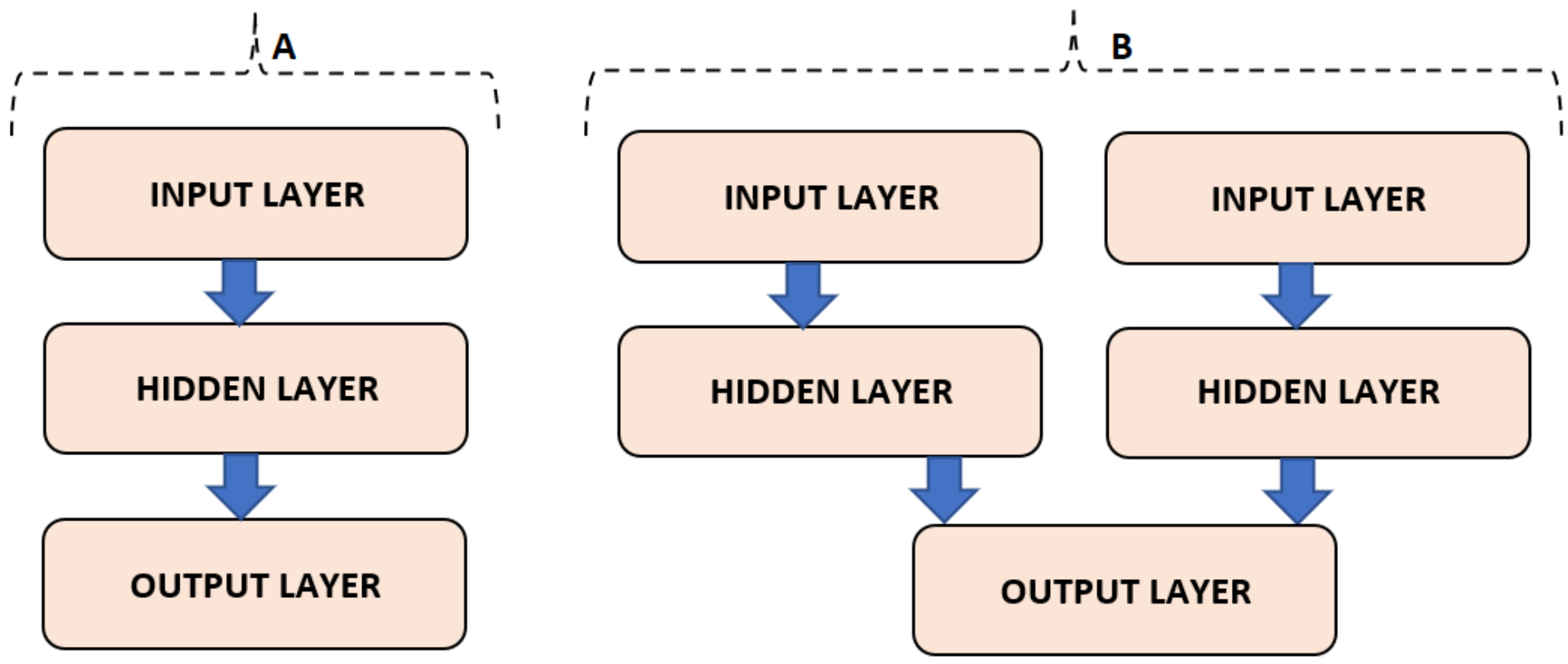

**Figure 1.** A schematic representation of two examples of the most commonly used networks/architectures - (A) CNN and (B) multi-stream CNN - for automated medical image analysis. Each block contains relevant layer nodes while relevant layer connections are generalized by a blue arrow symbol.

For example, in [26], the author focused on different values of the regularization hyperparameters to evaluate the effects such values had on the quality of semantic segmentation with U-net against GT segmentation. Regarding tunning other training hyperparameters, the author adopted a strategy of manually doing new adjustments only when 10% of all epochs were reached before achieving the 90% validation accuracy. Comparison of semantic segmentation with U-net against GT segmentation results demonstrated that the small value of L2 regularization could get semantic segmentation with U-net results much closer to ground truth segmentation results. However, the effects of such a regularization hyperparameter on fully automated quantification of MI were not studied in [26]. Therefore, in [27], the author presented the preliminary work related to fully automating the quantification of the MI. Here, the author chose the regularization hyperparameter value considering or following recommendations given in [26]. In [27], the quantification algorithm known as EWA, incorporated in the Segment CMR Software, quantified the infarct scar sizes during the process of full automation of the quantification of MI. EWA was based on expectation-maximization and a weighted intensity and in [28], the authors proved that it might serve as a clinical standard for quantification of MI in LGE-MRI images. Normally, quantification algorithms were applied to segmented structures to extract the essential diagnostic information such as shape, size, texture, angle, and motion [16]. Because the types of measurement and tissue vary considerably, numerous quantification techniques, including EWA, that addressed specific applications, were developed [16],[28]. In the preliminary work presented in [27], the author demonstrated that more than 50 % of the average infarct scar volume, 75% of infarct scar percentage, and 65 % of microvascular obstruction (mo) percentage were achieved with the EWA algorithm. However, in both previous works, [26] and [27], the effects of the size of LGE-MRI images in the training datasets on the deep learning training outcome or output of deep learning algorithms were not studied. Therefore, in this paper, the author studied such effects using different interpolation algorithms. To the best of the author's knowledge, image interpolation algorithms are divided into two major categories of non-extra-pixel and extra-pixel interpolation algorithms [34]. Unlike, the extra-pixel approach, the non-extra-pixel approach only uses original or source image pixels to produce or output interpolated images of the desired size[36]. Selected examples of such approaches-based interpolation algorithms are provided in part 2, sub-section 2.2. Given that the non-extra pixel category algorithm, such as nearest neighbor interpolation, is routinely used to interpolate ground truth masks due to its inherent advantage of not creating non-original or extra class labels in the interpolated masks (during the datasets image resizing processes), in this work, the author demonstrated the possibility and importance to improve the deep learning- based segmentation and MI quantification results by resizing images, in the training datasets, using extra pixel approach-based interpolation algorithms. In brief, the author first determined the new size of LGE-MRI images, of the reference training datasets, using extra-pixel approach-based interpolation algorithms and corrected errors or removed extra class labels in interpolated ground truth segmentation masks using a novel strategy developed for interpolation masks handling purposes. In this way, the author was able to evaluate how the change-in-image-size improves or worsens predictive capability or performance of deep learning-based U-net via semantic segmentation and quantification operations. It is important to note that, in this context, the U-net is used as (an existing and well documented) method to carry out deep learning-based semantic segmentation operations. It is also important to note that the nearest neighbor image interpolation algorithm normally produces heavy visual texture and edge artefacts that reduce or worsen the quality of interpolated images.

Fully automated quantification of the MI was achieved by the EWA algorithm applied to the outcome of automatic semantic segmentation with U-net**.** During experiments, common class metrics were used to evaluate the quality of semantic segmentation with U-net against the GT segmentation. And, arbitrary threshold, comparison of the sums, and sums of differences were used as criteria or options to estimate the relationship between semi-automatic and fully automated quantification of MI results. After experimental simulations, a close relationship between semi-automatic and fully automated quantification of MI results was more detected or identified in the case involving the dataset of bigger LGE MRI images than in that of the dataset of smaller LGE-MRI images.

In the next parts of this paper, the word *manual* may refer to semi-automatic or medical experts-based results while the word *automated* refers to fully automated or U-net-based results. The rest of the paper is organized as follows: Part II presents the materials and methods used to demonstrate effects. Part III presents a description of the dataset used, metrics, methods, U-net settings, and graphic card information. Part IV presents discussions related to the experimental results. Part V gives the conclusion of this work.

## 2. Materials and Methods

### *2.1. U-net Architecture*

U-Net is a CCN architecture widely used for semantic segmentation tasks [10]. It features a U-shaped design, comprising contracting and expansive paths.

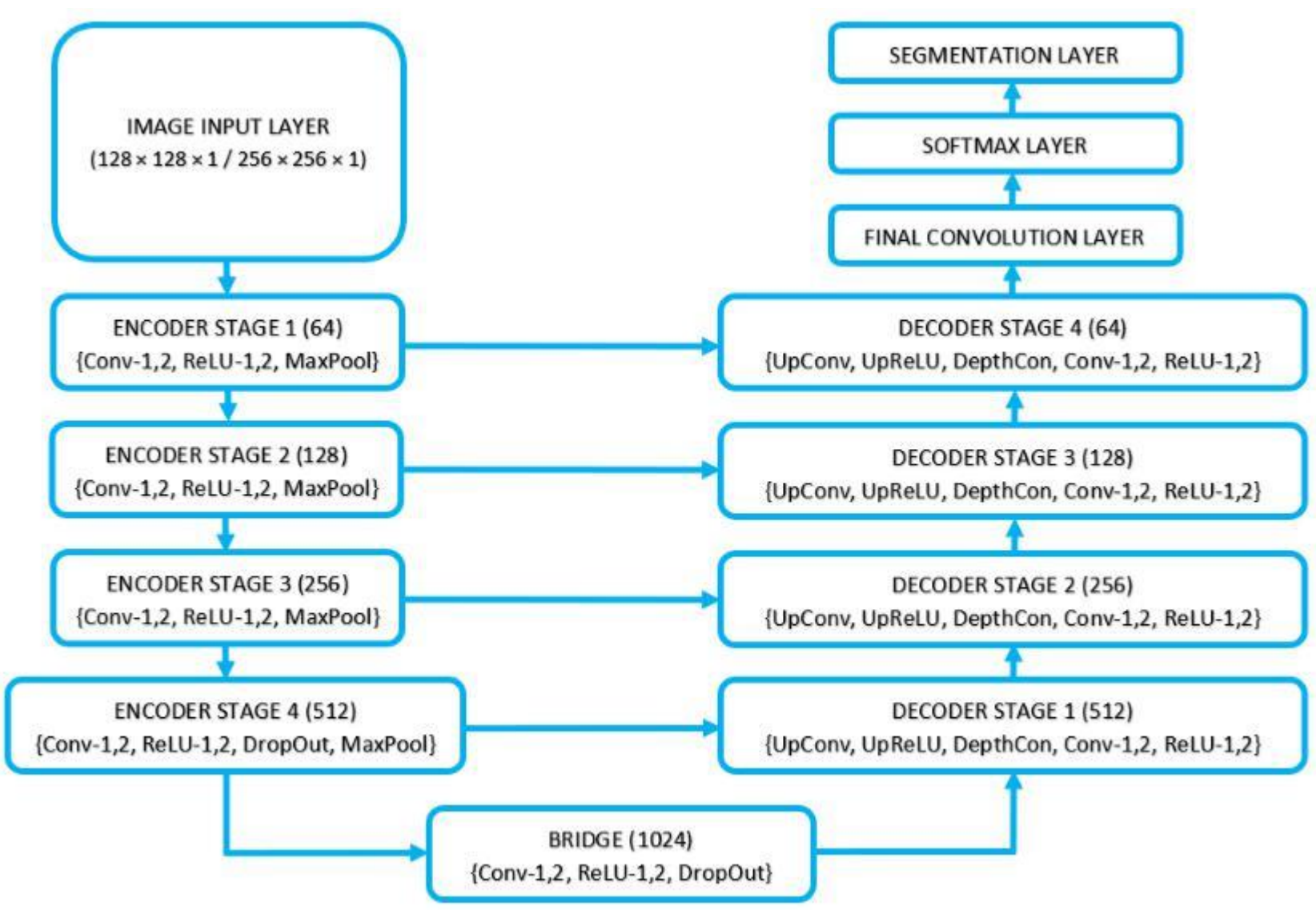

**Figure 2.** U-net architecture. *Conv* means convolution. *ReLU* is rectified linear unit. *DepthConv* is depth concatenation. *UpConv* means up-convolution or transposed convolution. *MaxPool* is Max Pooling.

In our experiments, we used the U-Net Layers function in MATLAB to easily create a U-Net architecture for semantic segmentation. This function follows the U-shaped architecture described in the original U-Net paper [10]. The contracting path consists of repeating blocks of convolution, ReLU activation, and max pooling. The expansive path involves transposed convolution, ReLU activation, concatenation with the downsampled feature map, and additional convolution. The *U-Net Layers function* provides options to customize the network, but note that it is just one implementation of the U-Net architecture. For more information, refer to the MATLAB documentation [51] and [52]. Figure 2 briefly shows the input and output layers, as well as the intermediate layers and connections, of a deep learning network as visualized by the *analyzeNetwork* function in MATLAB.

### *2.2. Selected Methods for Image Interpolation*

Interpolation is a technique that pervades or penetrates many applications [29], [30], [31], [32], [33], [34]. Interpolation is rarely the goal (in itself), yet it affects both the desired results and the ways to obtain them [16]. In this work, the nearest neighbor, bicubic, and Lanczos interpolation algorithms are used to determine the new size of LGE-MRI images in the training datasets, due to their acceptable performance and popularity in image processing and analysis software [35], [36], [37], [38], [39].

#### 2.2.1. Nearest Neighbor Interpolation

Nearest neighbor interpolation (NN) is the fastest image interpolation method that belongs to the non-extra pixel category [35], [36], [38]. NN does not include a weighted weighting function, instead, it is based on the (linear scaling and) rounding functions that decide which pixel to copy from source to destination image [35], [36], [38].

#### 2.2.2. Bicubic Interpolation

Bicubic interpolation (BIC) is an extension of cubic interpolation for interpolating data points on a two-dimensional regular grid that belongs to the extra pixel category [36], [37]. BIC uses a weighted average of 16 samples to achieve the interpolated value of the new pixel sample [37].

#### 2.2.3. Lanczos3 Interpolation

Lanczos interpolation (LCZ) is based on the 3-lobed Lanczos window function as the interpolation function [39], [40]. LCZ also belongs to the extra pixel category [36]. LCZ uses source image pixels (36 pixels) and interpolates some pixels along the x-axis and y-axis to produce intermediate results [39], [40].

### *2.3. Histogram Visualization of Interpolated GT Segmentation Masks*

After changing the size of LGE images in the reference dataset or simply after interpolating LGE-MRI images and GT segmentation images in the training dataset, there comes a risk of misplaced class labels in the interpolated GT segmentation masks, or extra classes or class labels are created in the mask regions where they should not be present. To visualize and examine possible extra class labels after GT segmentation masks interpolation, the histogram visualization technique is used, and histograms of interpolated GT segmentation masks are presented in Figure 3. Figure 3-(top-left) shows the histogram of the non-interpolated GT mask of the size 128 × 128. Figure 3-(top-right) shows the histogram of the NN interpolated GT mask of the size 256 × 256. In both Figure 3-(top-left) and (top-right) cases, the histograms look the same way. Both histograms show three classes regardless of how images are obtained. In that case, the NN interpolation did not change the number of classes of the original GT segmentation mask - and the reason was that the NN did not create extra pixels in the interpolated GT segmentation masks [36]. Figure 3-(bottom-left) and Figure 3-(bottom-right) show histograms of the BIC and LCZ interpolated GT segmentation images, respectively. As can be seen, in both Figure 3-(bottom-left) and (bottom-right) cases, the histograms do not look the same way. On top of that, the histograms show more than three classes (instead of the expected three classes).

In Figure 3-(bottom-left) and (bottom-right) cases, BIC and LCZ interpolation algorithms changed the number of classes of the original GT segmentation mask, thus requiring removing extra class labels to keep the original number of classes unchanged. Note that, due to NN interpolation artefacts, reducing the quality of interpolated images, it was necessary to also use other interpolation algorithms (even if doing that re-quired extra effort).

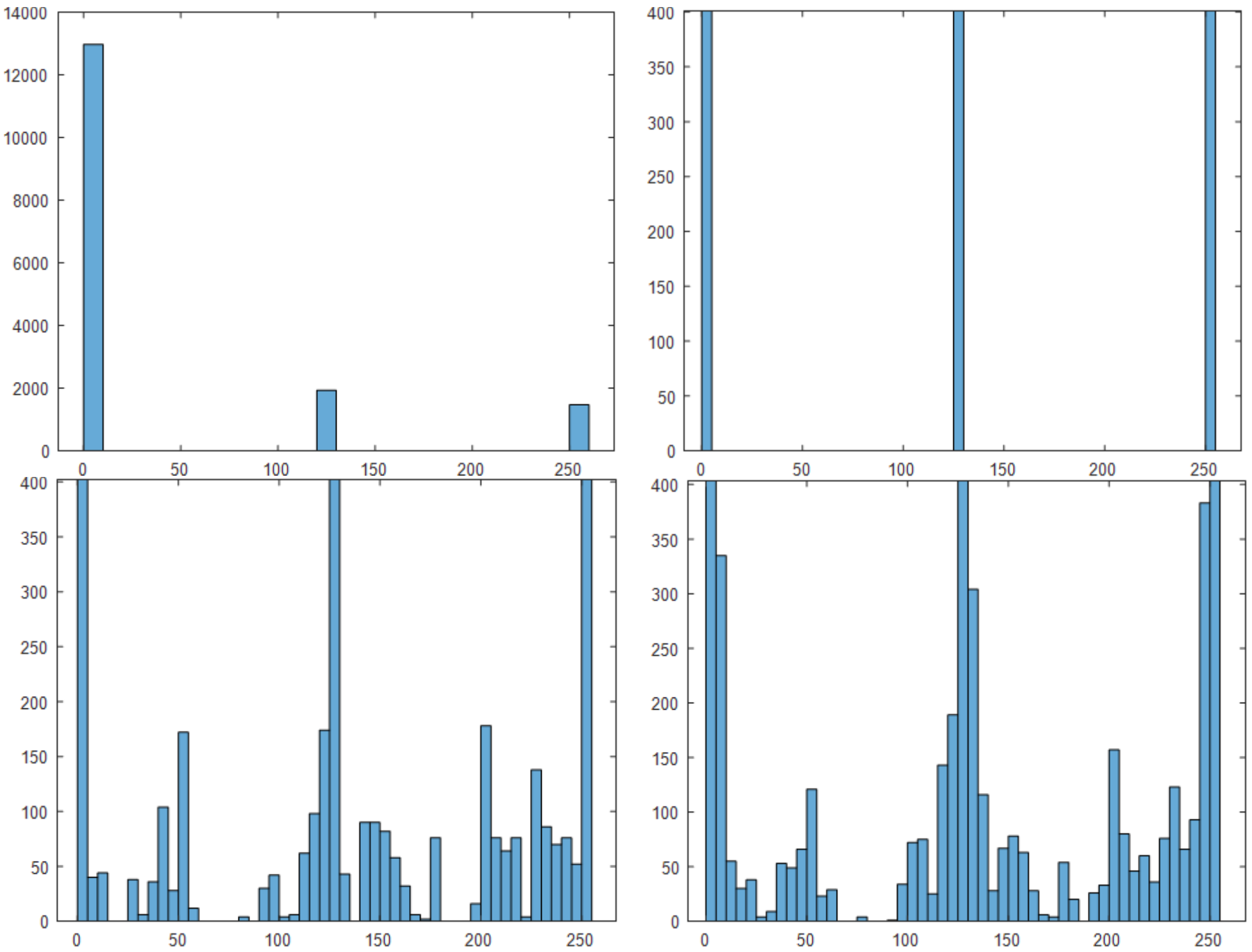

**Figure 3.** Histograms: Top left: GT segmentation mask of the size 128 × 128. Top-right: NN-based GT segmentation mask of the size 256 × 256. Bottom-left: BIC-based GT segmentation mask of the size 256 × 256. Bottom-right: LCZ-based GT segmentation mask of the size 256 × 256.

### *2.4. A Novel Strategy for Removing Extra Class Labels in Interpolated GT Segmentation Mask*

First, it is important to remind that the nearest neighbor interpolation would be the simplest option to interpolate GT masks due to its inherent advantage of not creating non-original or extra class labels in the interpolated masks. The only problem is the deterministic rounding function on which its pixel selection strategy is based [53]. Such a strategy slightly shifts the entire image content to some extent and is responsible for creating heavy jagged artefacts in interpolation results [36], [38], [53]. Also, it is important to remind that extra-pixel category-based interpolation algorithms do not shift the image content and do not produce heavy jagged artefacts. The only problem is that their weighting functions create extra class labels once used to interpolate GT masks.

There are certainly many strategies, one can think of, to remove extra class labels thus solving an image processing problem of this kind. For example, it could be easier to think or imagine that extra class labels could only be removed using a function based on Equation 1 or closely related.

$$T(x) = \begin{cases} 0 & x < 64 \\ 255 & 192 < x \\ 128 & otherwise \end{cases} \tag{1}$$

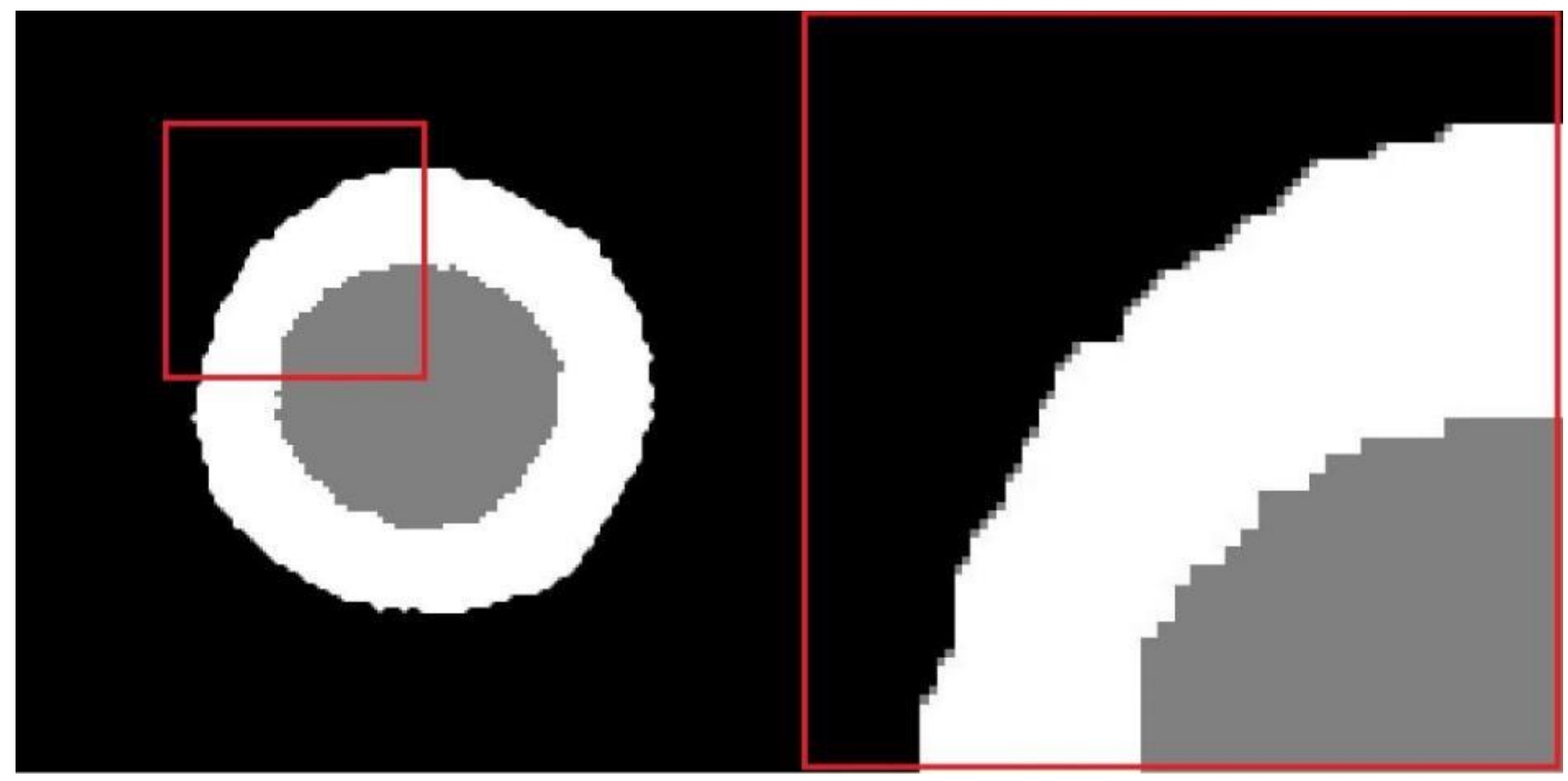

**Figure 4.** Example showing the bicubic (BIC) interpolated GT segmentation mask after removing extra class labels using the Equation 1-based function.

Figure 4 shows the outcome of implementing Equation 1's function to remove extra class labels in the interpolated GT segmentation mask. As can be seen, the Equation 1's idea did not work as one would expect – because, around edges between the class represented by 0- and 255-pixel labels, there were still pixel labels that looked like 128-pixel labels, which should not be the case. Another strategy, which is routinely used, is the use of extra-pixel-category-based algorithms for training images and the nearest neighbor interpolation algorithm for training masks. To the best of the author's knowledge, that is not a better strategy due to the risk of misalignment, of both endocardium and epicardium outlines in nearest neighbor interpolated GT masks, which is likely to worsen the annotation errors thus negatively affecting the accuracy of segmentation with deep learning methods.

Therefore, the author developed a better (and dedicated) strategy focusing on removing extra class labels in interpolated GT images and the developed strategy is based on three important techniques/operations, namely (1) thresholding, (2) median-filtering, and (3) subtraction. In this way, extra class labels are removed in five steps (designated by the S letter), as shown in Figure 5.

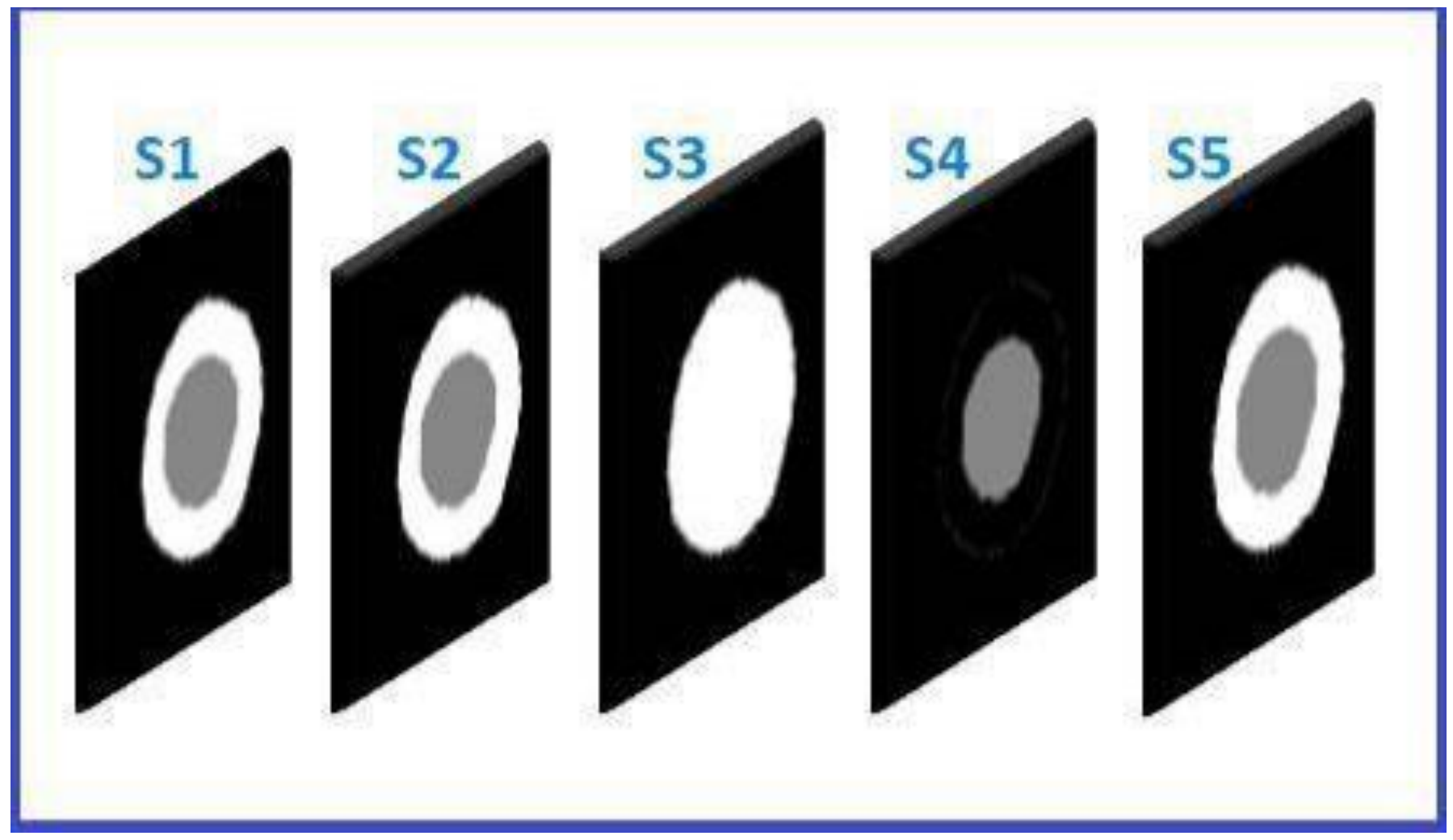

**Figure 5.** Five steps to remove extra class labels in BIC interpolated GT seg-mentation masks.

**Step 1**: Initially, a GT segmentation mask size is resized to the size of interest using either BIC or any other extra pixel approach-based image interpolation algorithms. Here, the resulting mask is referred to as S1 and is shown in Figure 6 (a). Note that S1 is a mask to filter or in which extra class labels must be removed.

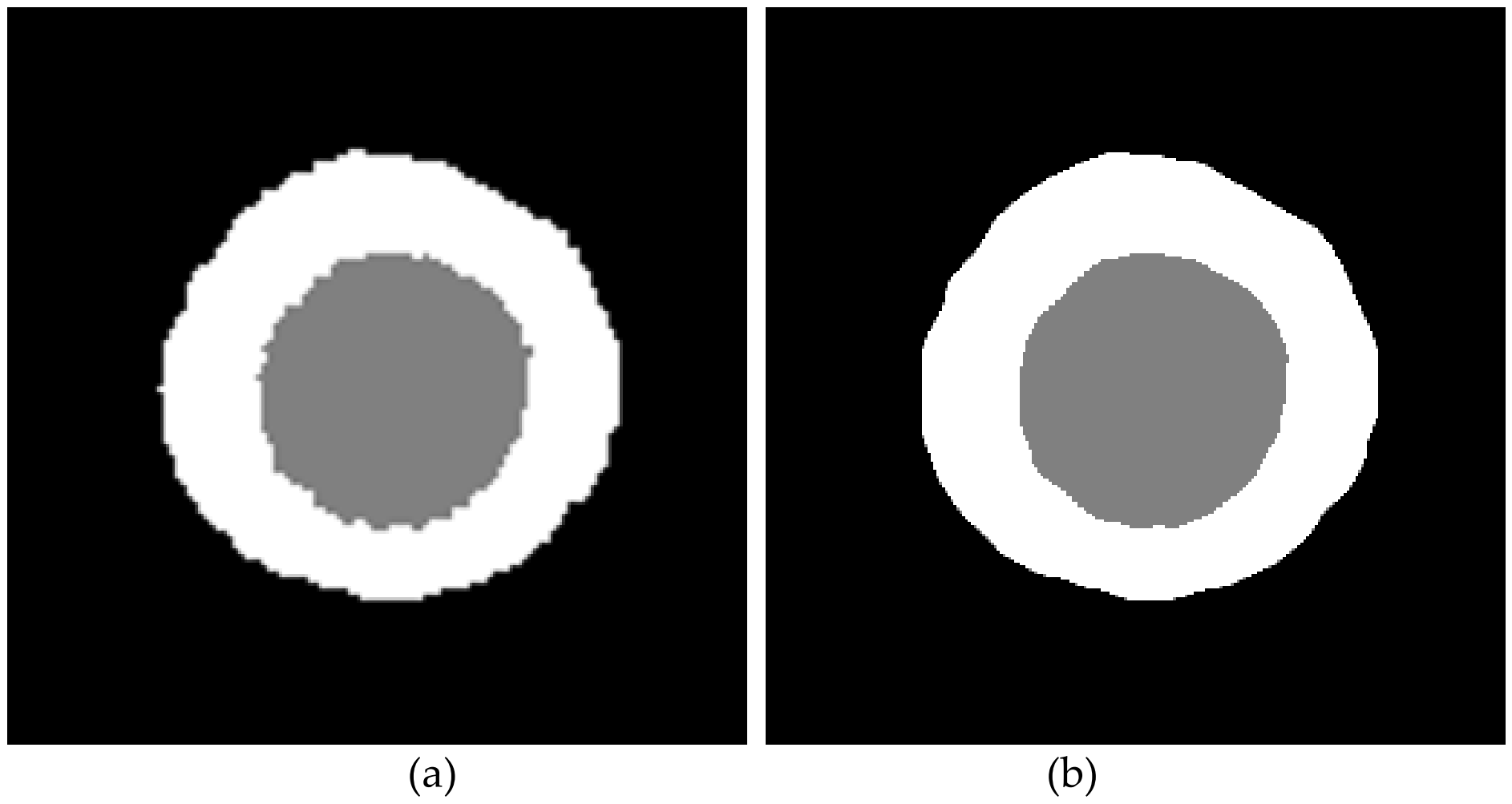

(a) (b)

**Figure 6**. (a) S1 and (b) S2 output images of the size 256 x 256

**Step 2**: Extra class labels of S1 falling outside the desired class labels range are removed via thresholding. The resulting mask is referred to as S2 as shown in Figure 6 (b). But there are still few extra labels of pixels that remained scattered on the S2 surface (e.g., see Figure 6 (b)) that are still present even after applying the median filter.

**Step 3**: Unwanted class labels of S2 (e.g., 128) are removed and the result is referred to as S3 as shown in Figure 7 (a).

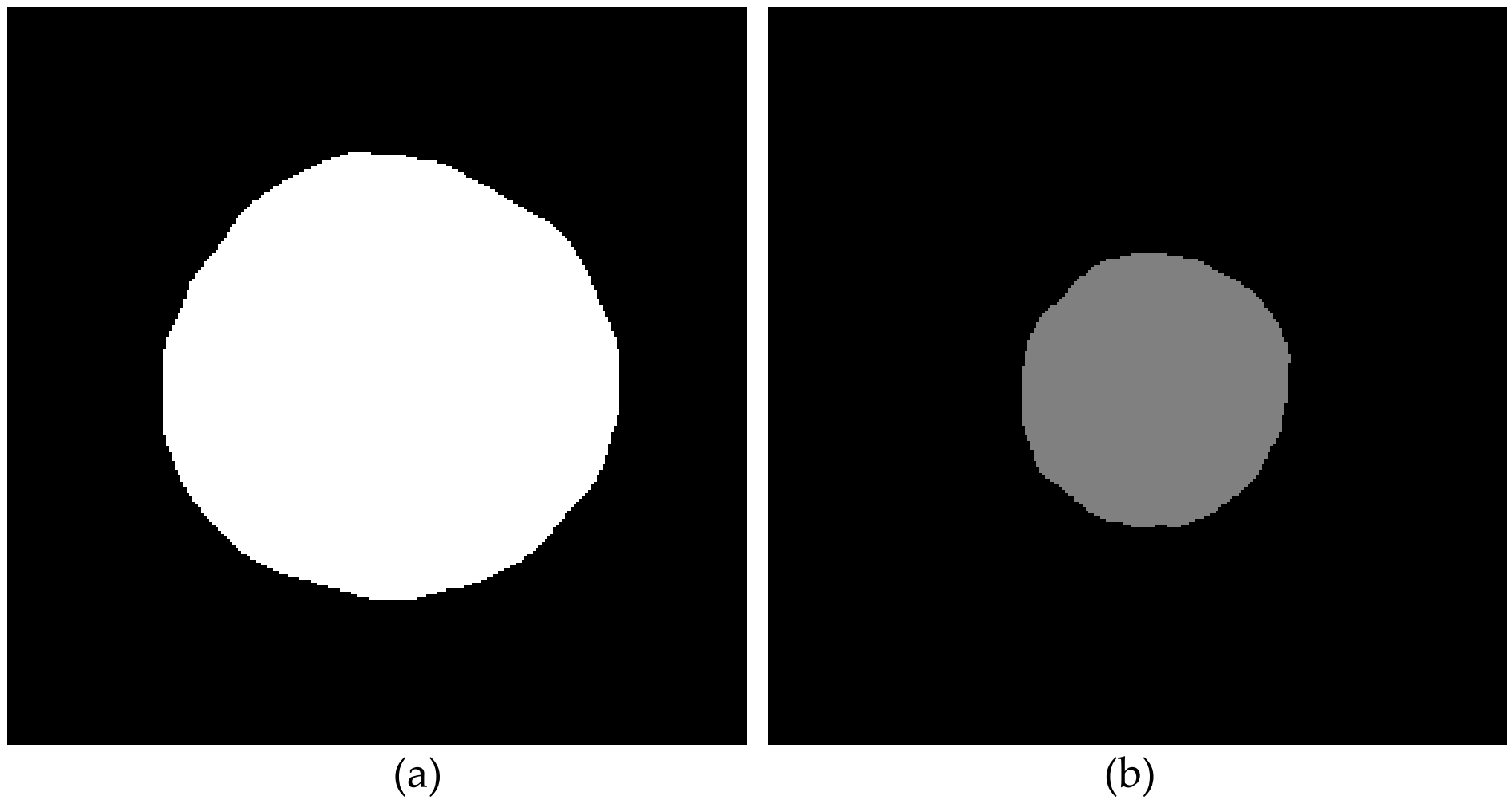

(a) (b)

**Figure 7**. (a) S3 and (b) S4 output images of the size 256 x 256

**Step 4**: Again, other unwanted class labels of S2 (e.g., 255) are removed and the result is referred to as S4 as shown in Figure 7 (b). Here, it is important to note that after excluding class labels (255) there were still class labels (128) on the epicardium outline there are still present but removed using the median filter.

**Step 5**: Here, S4 is subtracted from S3 only when any class label of S3 is equal to 0 (this is to be done to avoid adding one to zero pixels). When none of the class labels of S3 is equal to 0, S4 is subtracted from S3, and one is added to the difference (because in that case, the difference is equal to 127). Figure 8 (a) shows the output mask, referred to as S5 and Figure 8 (b) shows the input mask whose interpolated version was shown as S1. Note that all these five steps are executed in one single operation.

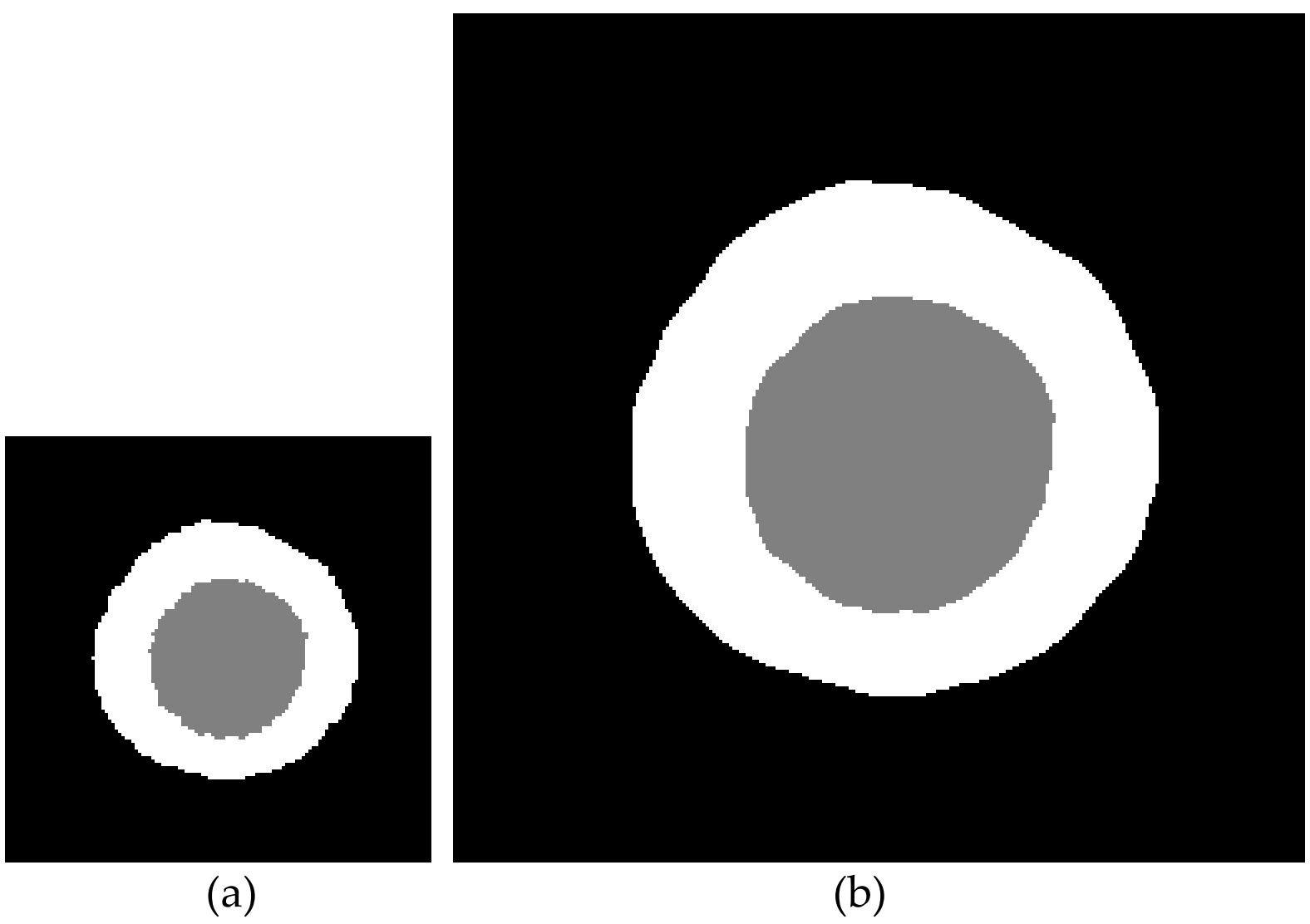

**Figure 8**. (a) input mask of the size 128 x 128. (b) S5 output mask of the size 256 x 256

## 3. Results

The description of the dataset, metrics, methods, U-net hyperparameter settings, and graphic card information is provided in this part. However, details on experimental results are provided in combination with discussions, in the discussion part.

### 3.1. Image datasets

The reference dataset included a total of 3587 LGE MRI images and GT segmentation masks of the size 128 × 128. GT segmentation masks were converted from semi-automatically annotated LGE-MRI images using the Segment CMR Software tool-version 3.1.R8225 [41] Each GT segmentation mask consisted of three classes, with class IDs, corresponding to 255-, 128-, and 0-pixel labels. As done in [26] and [27], the main dataset was split into three datasets, namely: the training set (60% of the main dataset), the validation set (20% of the main dataset), and the test set (20% of the main dataset). Note that information or details related to clinical trial registration can be found or are provided in [28], therefore are not included in this section.

### 3.2. Metrics and Methods

To evaluate the quality of the masks from semantic segmentation using U-net against the GT segmentation, class metrics, namely: classification accuracy, intersection over union (IoU), and mean (boundary F-1) BF score were used to (1) estimate the percentage of correctly identified pixels for each class, (2) achieve statistical accuracy measurement that penalizes false positives and (3) see how well the predicted boundary of each class aligns with the true boundary or simply use a metric that tends to correlate with human qualitative assessment, respectively [42], [43]. In addition, Sørensen-Dice similarity coefficients were used to evaluate the quality of U-nets' segmented output masks against GT segmentation output masks. To evaluate the relationship between semi-automatic or medical experts-based and fully automated quantification of MI results, the values or sizes of the infarct scar volume and percentage, as well as the microvascular obstruction percentage were calculated or obtained by applying the EWA algorithm on automatically segmented masks [26], [27], [28]. It is important to also mention that the simulation software was MATLAB R2020b. Segment CMR software worked well with MATLAB R2019b.

### 3.3. U-net settings and graphic cards

The training hyperparameters were manually adjusted based on the observation of the training graph, with the possibility for new adjustments when 10% of all epochs were reached before the training accuracy reached 90% [26]. Here, U-net's training hyperparameters, manually adjusted, included the number of the epochs = 180, minimum batch size = 16, initial learning rate = 0.0001, L2 regularization = 0.000005 (referring to recommendations provided in [26]). Adam was the optimizer. The loss function used in this case was the default cross-entropy function provided by the U-Net Layers function. Further information on this function can be found in reference [52]. The execution environment was multi-GPU with both Nvidia Titan RTX and Nvidia GeForce RTX 3090 graphic cards. Data augmentation options used to increase the number of images in the dataset used to train the U-net were a random reflection in the left-right direction as well as the range of vertical and horizontal translations on the interval ranging from -10 to 10.

## 4. Discussion

### *4.1. Evaluation of the effects of image size on the quality of automatic segmentation with U-net against the GT segmentation*

In the effort to evaluate the effects of image size on the quality of deep learning-based segmentation (or deep learning performance or outcome on segmentation), when the image size is changed from 128 × 128 to 256 × 256, three classes or regions of segmented masks are evaluated using Accuracy, IoU, mean BF score. Before going into the evaluation of each region, it is important to note that C128 represents the U-net trained on LGE-MRI images of the size 128 × 128. N256F, B256F, and L256F represent the U-nets trained on LGE-MRI images of size 256 × 256 obtained after doing interpolation operations using the NN, BIC, and LCZ methods and filtering the corresponding GT segmentation masks using the strategy introduced in Part II. N256U, B256U, and L256U also represent the U-nets trained on LGE-MRI images of the size 256 × 256 obtained after doing interpolation operations using the NN, BIC and LCZ methods but without removing extra class labels.

#### 4.1.1. Region 1

Region 1 represents the class of the GT segmentation mask corresponding to the 255-pixel label. Class metrics-based results from automated segmentation with U-net of this region are shown/provided in Figure 9. As can be seen, in Figure 9, N256F and N256U produced the same results in terms of Accuracy, IoU, and mean BFScore, thus confirming the no need for filtering the NN interpolated GT segmentation images. Also, as can be seen, the C128-based network led to the poorest performance among other networks compared to or mentioned in terms of Accuracy, IoU, and mean BFScore.

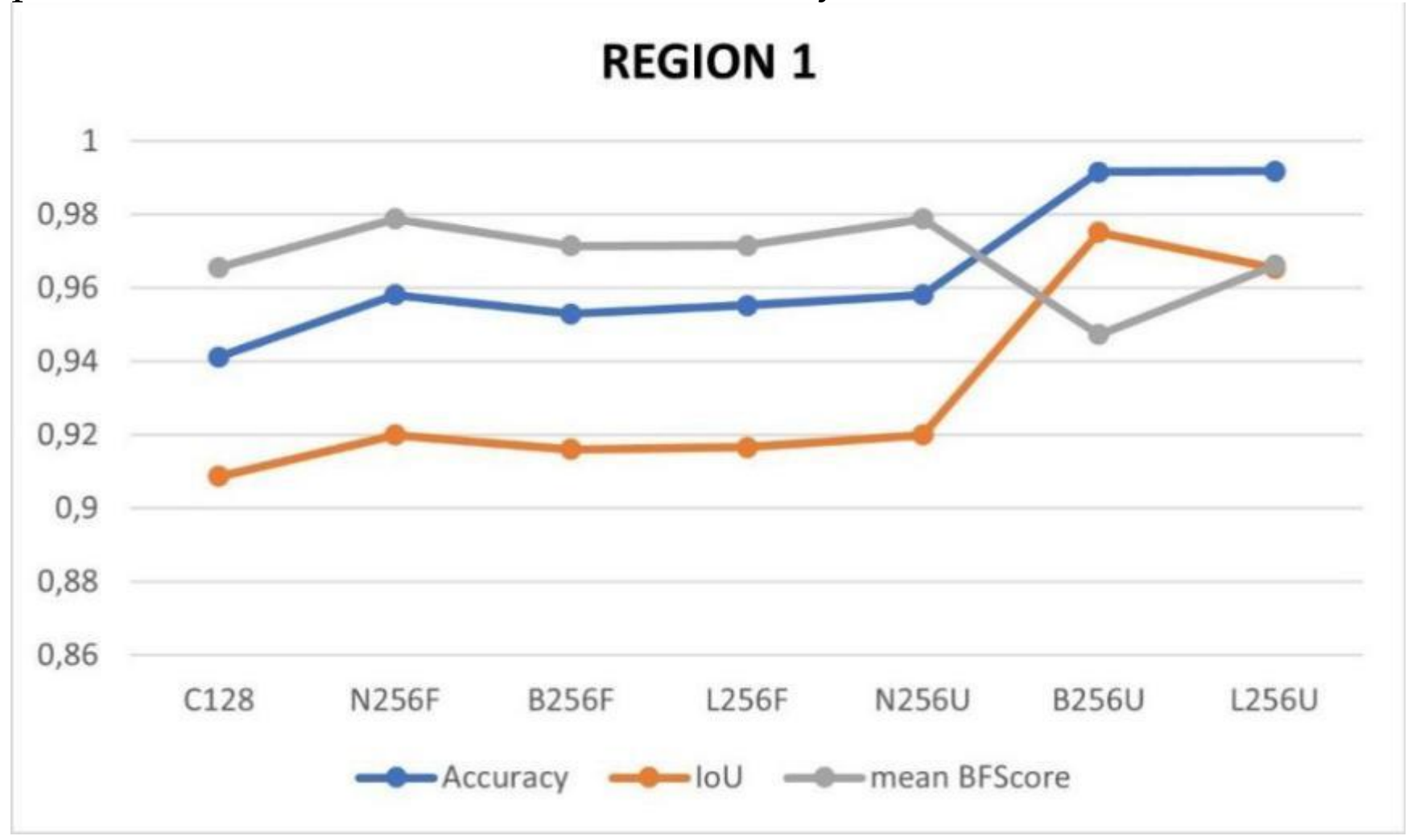

**Figure 9.** Segmentation results: Region 1.

#### 4.1.2. Region 2

Region 2 represents the class of the GT segmentation mask corresponding to the 128-pixel label. Class metrics-based results from automated segmentation with U-net of this region are shown in Figure 10.

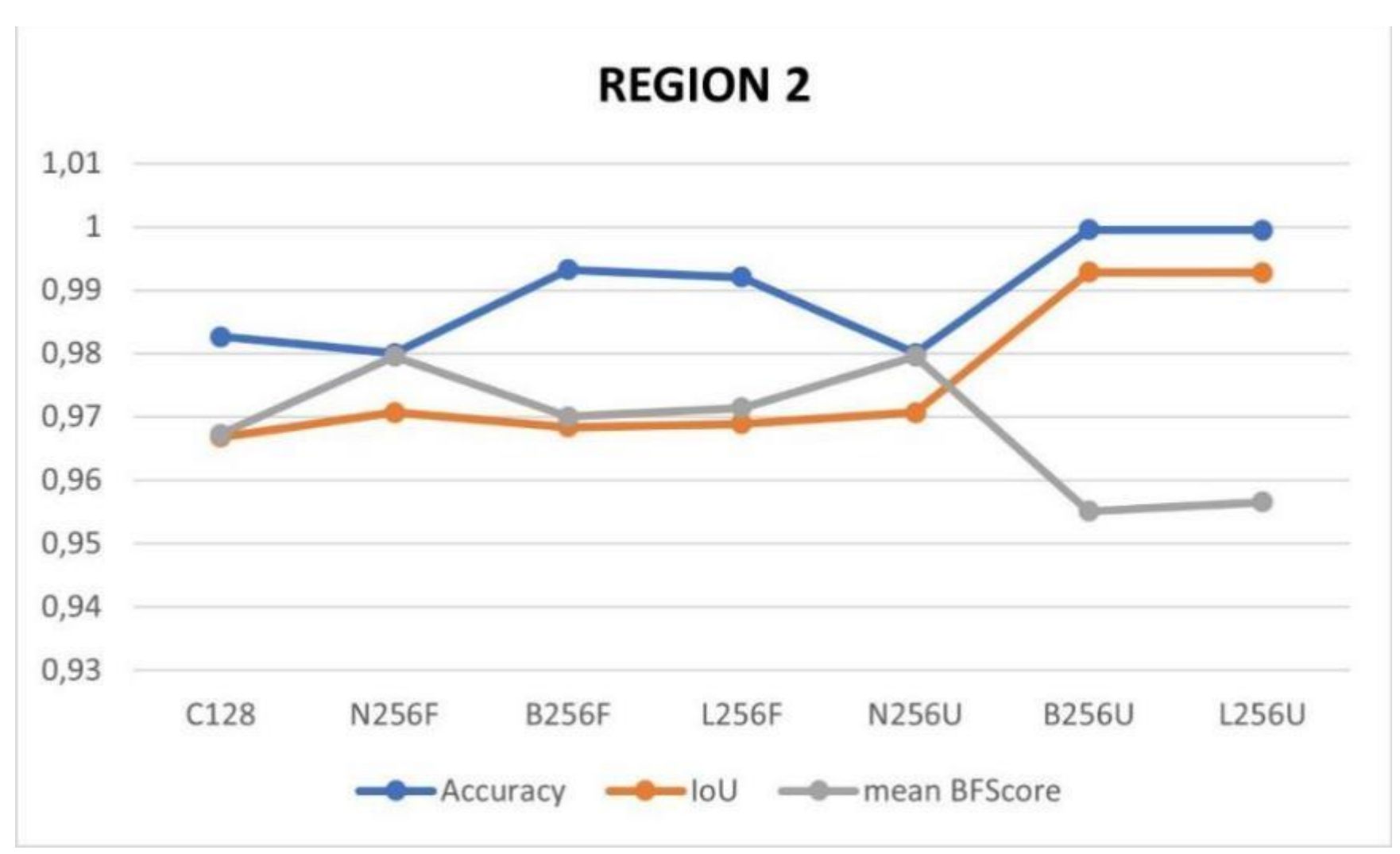

**Figure 10.** Segmentation results: Region 2.

Here, again, N256F and N256U produced the same results in terms of Accuracy, IoU, and mean BFScore, thus confirming again the no need for filtering the NN interpolated GT segmentation images. Here, C128 did not always achieve the poorest performance, among all other networks mentioned, because, as can be seen, in terms of mean BF score, C128 outperformed B256U and L256U. In terms of accuracy, C128 outperformed the N256F/U. Only, in terms of IoU, the C128-based network achieved the poorest performance.

4.1.3. Region 3

Region 3 represents the class of the GT segmentation mask corresponding to the 0-pixel label. Class metrics-based results from the automated segmentation with the U-net of this region are shown in Figure 11. As can be seen, for the third time that N256F and N256U produced the same results in terms of Accuracy, IoU, and mean BFScore, thus confirming the no need for filtering NN interpolated GT segmentation images. Again, the C128 did not always achieve the poorest performance among all other networks mentioned. For example, in terms of mean BF score, C128 outperformed L256U. In terms of accuracy, C128 outperformed the N256F/U, B256F, L256F, and L256U. Only, in terms of IoU, the C128-based network achieved the poorest performance.

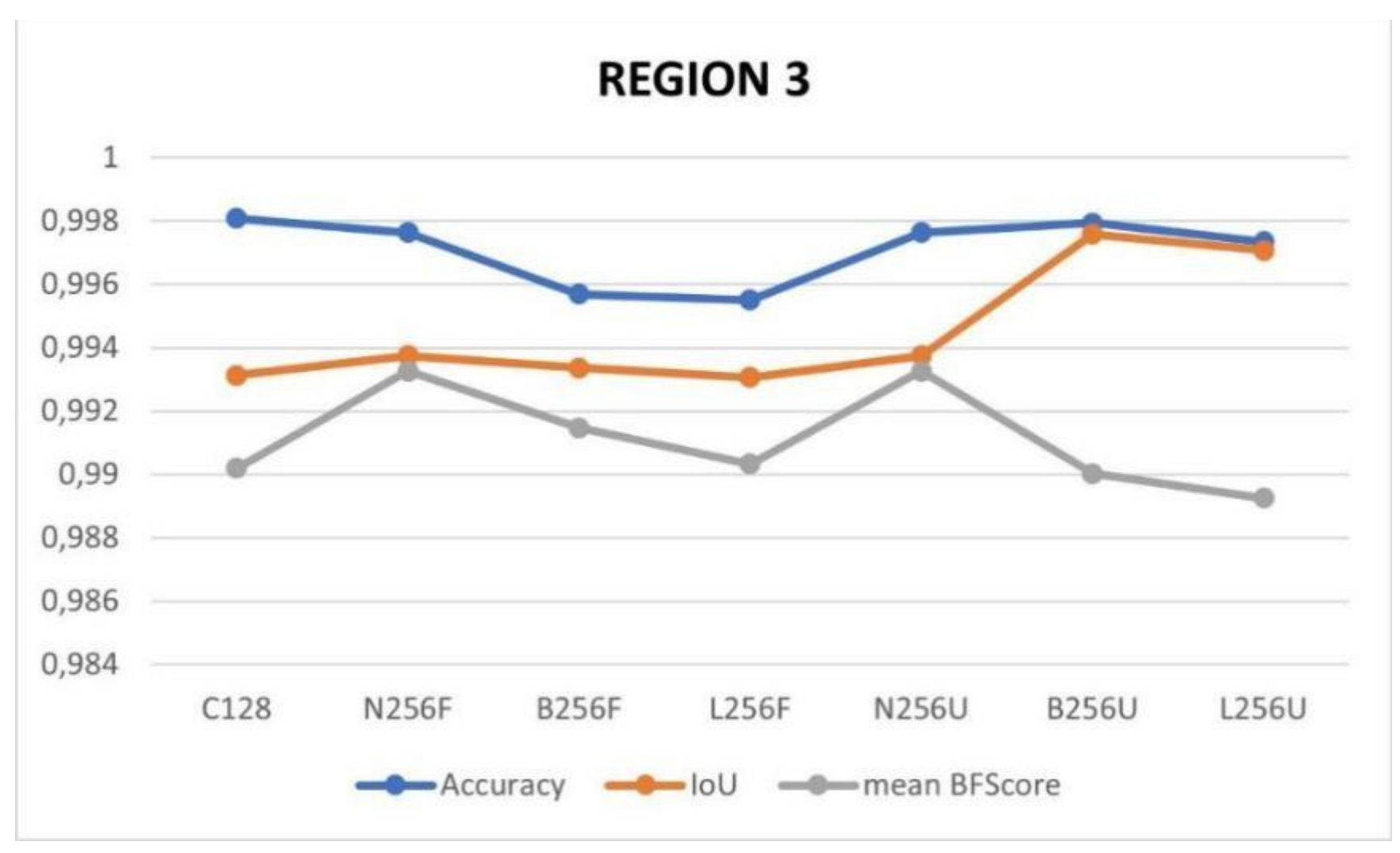

**Figure 11.** Segmentation results: Region 3.

4.1.4. Comparison of final validation and global accuracies of trained U-nets

Table 1 shows the final validation and global accuracies achieved by each U-net mentioned. Also, Table 1 shows that the validation and global accuracies achieved are generally in the same range, thus are no overfitting effects to be worried about. Note that previous experiments involving U-net-based segmentation demonstrated that filtering NN interpolated masks was not fruitful (see Figure 9, Figure 10, and Figure 11 as well as relevant discussions). In this regard, there is no more N256F or N256U but only N256, as shown in Table 1. Also, Table 1 shows that the training time of C128 is approximately half of the training time taken by other U-nets.

**Table 1.** U-net | Validation accuracy, Global accuracy, and Training time.

| Network | Validation accuracy | Global accuracy | Training time |
|---|---|---|---|
| C128 | 0.9908 | 0.99078 | 109 min 56 sec |
| N256 | 0.9919 | 0.9918 | 225 min 52 sec |
| B256F | 0.9914 | 0.99126 | 225 min 31 sec |
| L256F | 0.9916 | 0.9912 | 226 min 35 sec |
| B256U | 0.9977 | 0.99756 | 258 min 35 sec |
| L256U | 0.9973 | 0.9972 | 265 min 30 sec |

4.1.5. Performance evaluation of the U-net against Segnet

Segnet is another type of CNN designed for semantic image segmentation [46], [47]. To the best of the author's knowledge, these are the two that directly accept training sets of 2D grayscale images and whose source codes or functions are easily found for comparison purposes. In this section, the performance of Segnet is evaluated against the performance of U-net, and decisive performance results (in terms of Accuracy, IoU, and mean BFScore) are provided in Figure 12, Figure 13, Figure 14, and Table 2. Note that on these three Figures' y-axis, 0 to 3 or 3.5 are simply graphical scale values, automatically selected by MS Excel, and only represent how the real values differ from each other.

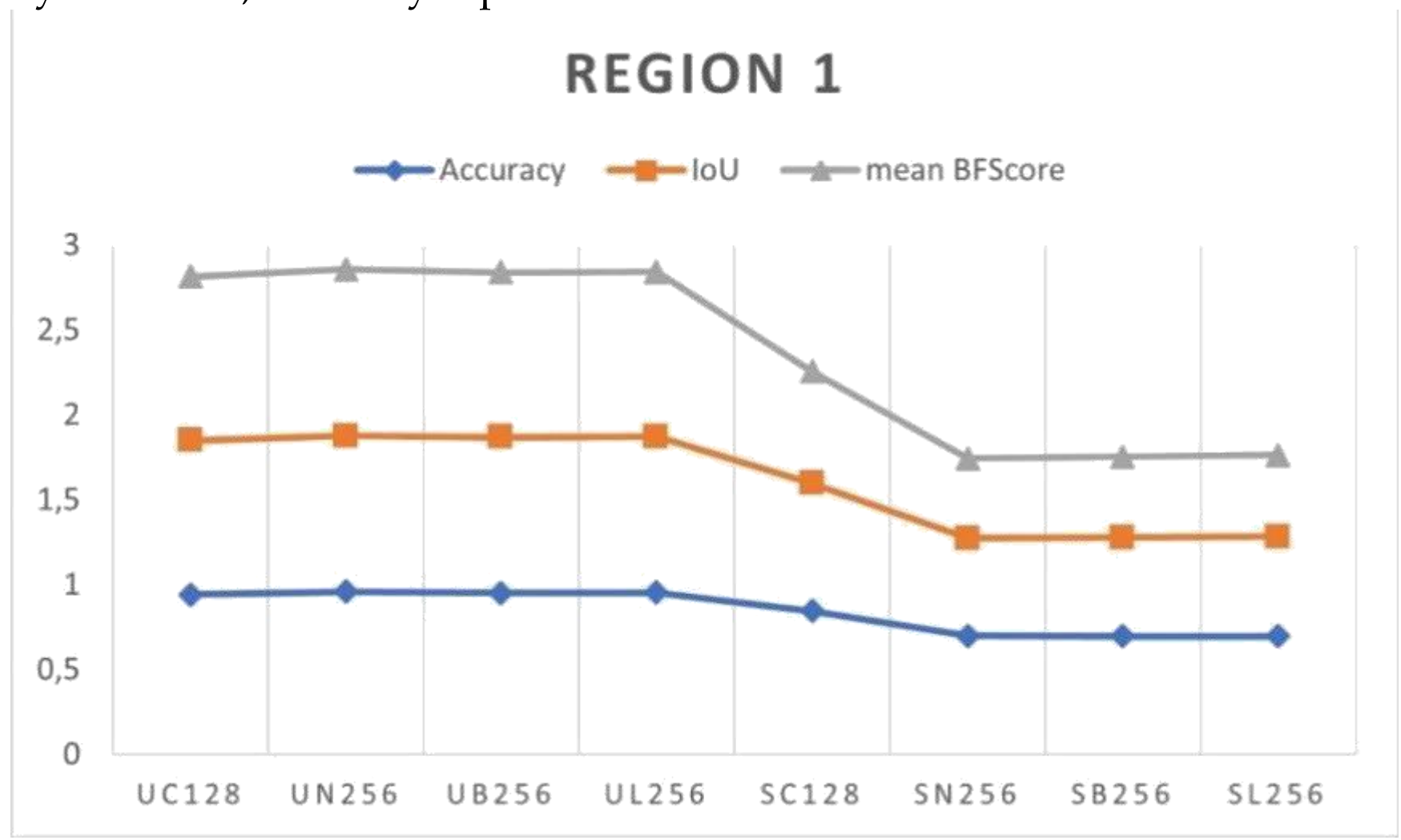

**Figure 12.** U-net vs Segnet | Segmentation Results | Region 1.

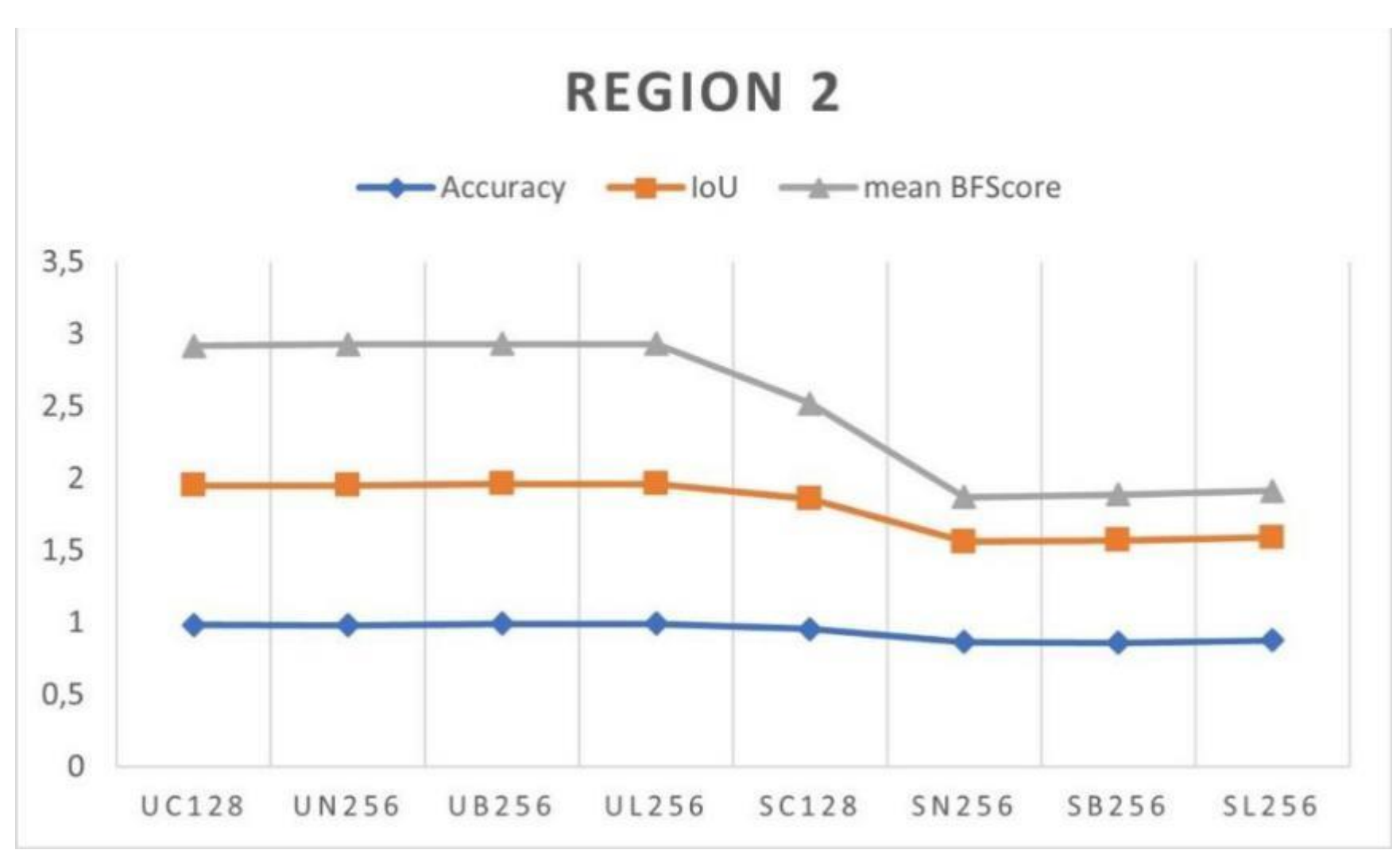

**Figure 13.** U-net vs Segnet | Segmentation Results | Region 2.

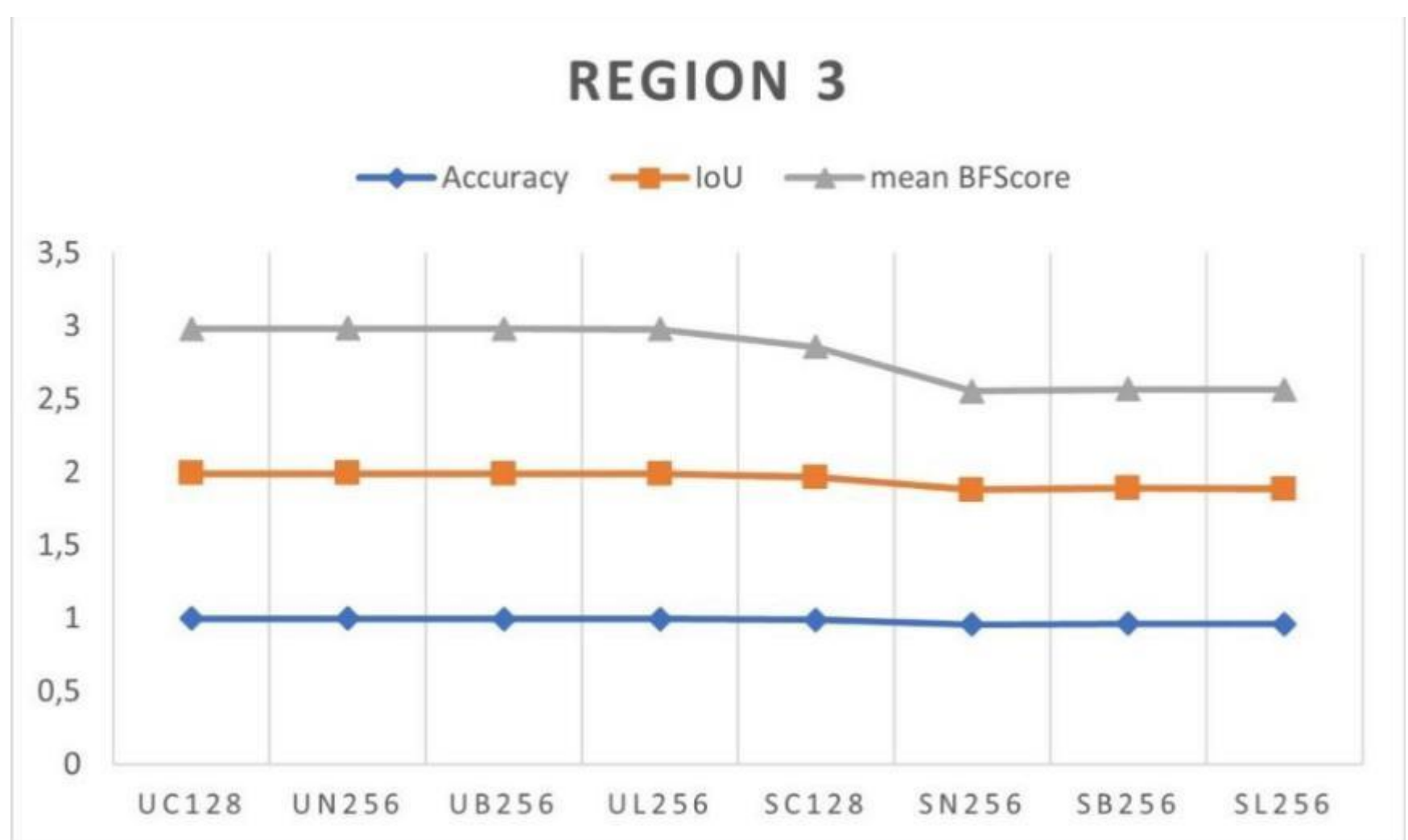

**Figure 14.** U-net vs Segnet | Segmentation Results | Region 3.

**Table 2.** U-net vs Segnet | Validation Accuracy, Global Accuracy, and Training Time.

| Network | Validation Accuracy | Global Accuracy | Training Time |
|---|---|---|---|
| UC128 | 0.9908 | 0.99078 | 109 min 56 sec |
| UN256 | 0.9919 | 0.9918 | 225 min 52 sec |
| UB256 | 0.9914 | 0.99126 | 225 min 31 sec |
| UL256 | 0.9916 | 0.9912 | 226 min 35 sec |
| SC128 | 0.9709 | 0.97149 | 144 min 08 sec |
| SN256 | 0.9221 | 0.92137 | 540 min 15 sec |
| SB256 | 0.9255 | 0.92499 | 548 min 42 sec |
| SL256 | 0.9244 | 0.92474 | 542 min 52 sec |

4.1.6. Evaluation of automated segmentation with U-net and GT segmentation using LGE MRI test images

From left to right, Figure 15, Figure 16, Figure 17, Figure 18, Figure 19, and Figure 20 show different columns of LGE-MRI test images and masks. Here, in each figure's case, the *first* column shows LGE-MRI test images. The *second* column shows GT segmentation masks. The *third* column shows segmented output masks using U-nets. The fourth column shows differences between GT segmentation masks and segmented output masks using U-nets. Such a difference is highlighted by colors. Here, it is important to note that the greenish and purplish regions highlight areas where the segmentation results differ from the GT segmentation mask. Also, note that dice indices are also provided in the caption of each figure in support of the qualitative evaluation. Comparing the dice indices in the caption of Figure 15 to those in the caption of Figure 16, it can be seen, that the C128-based network was outperformed only three times by the N256-based network. Next, C128 was outperformed three times by B256F (see Figure 17's caption), and four times by L256F (see Figure 18's caption). However, C128 was outperformed zero times by both B256U and L256F (see Figure 19-Figure 20's captions), therefore, U-nets based on unfiltered images were excluded from further discussions. Only, U-nets based on filtered images (previously labeled B256F and L256F) were kept and included in further discussions, as B256 and L256, respectively.

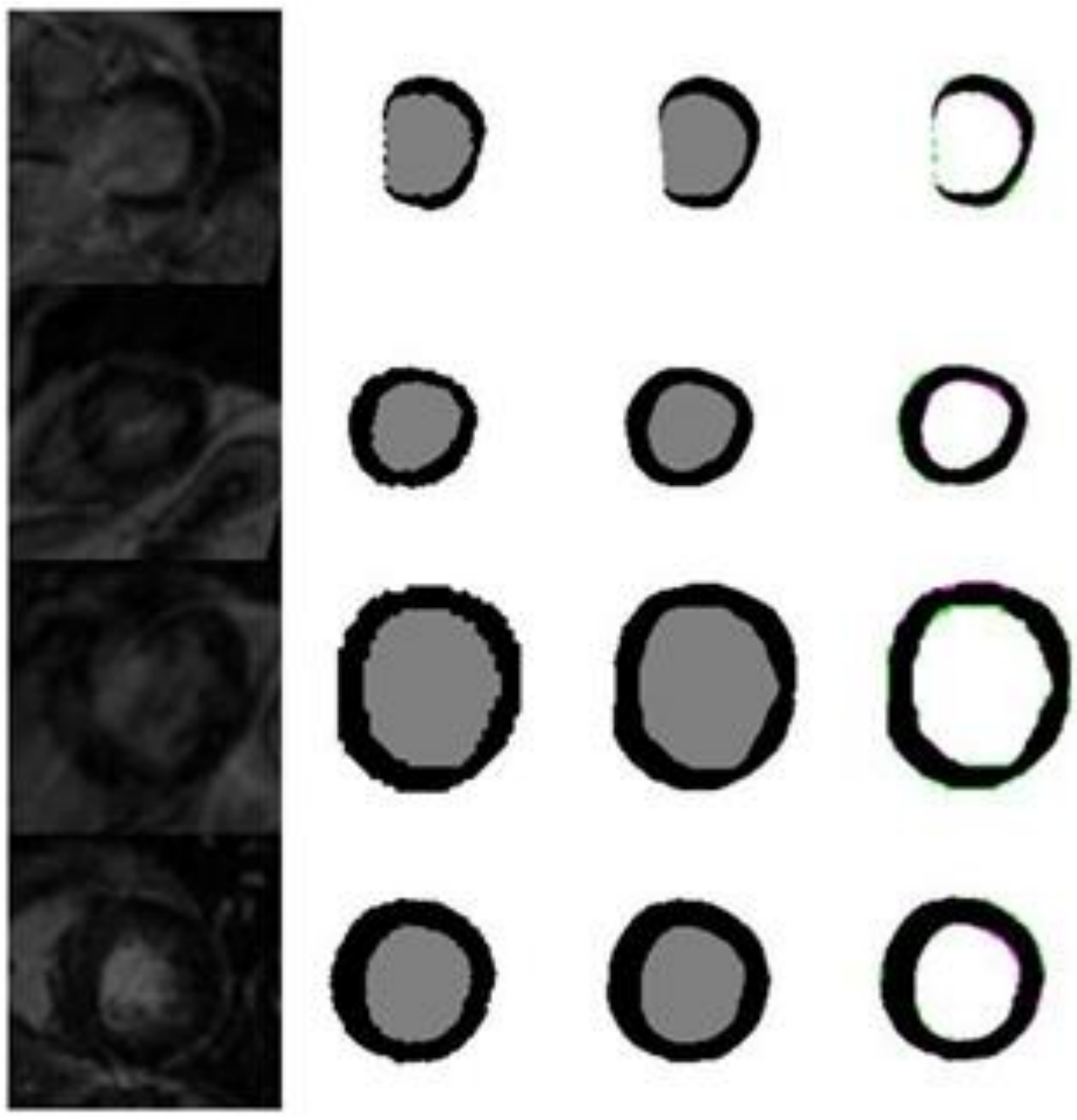

**Figure 15.** C128 segmented output masks | From top to bottom: Dice indices are equal to 0.9953, 0.9945, 0.9873, and 0.9929.

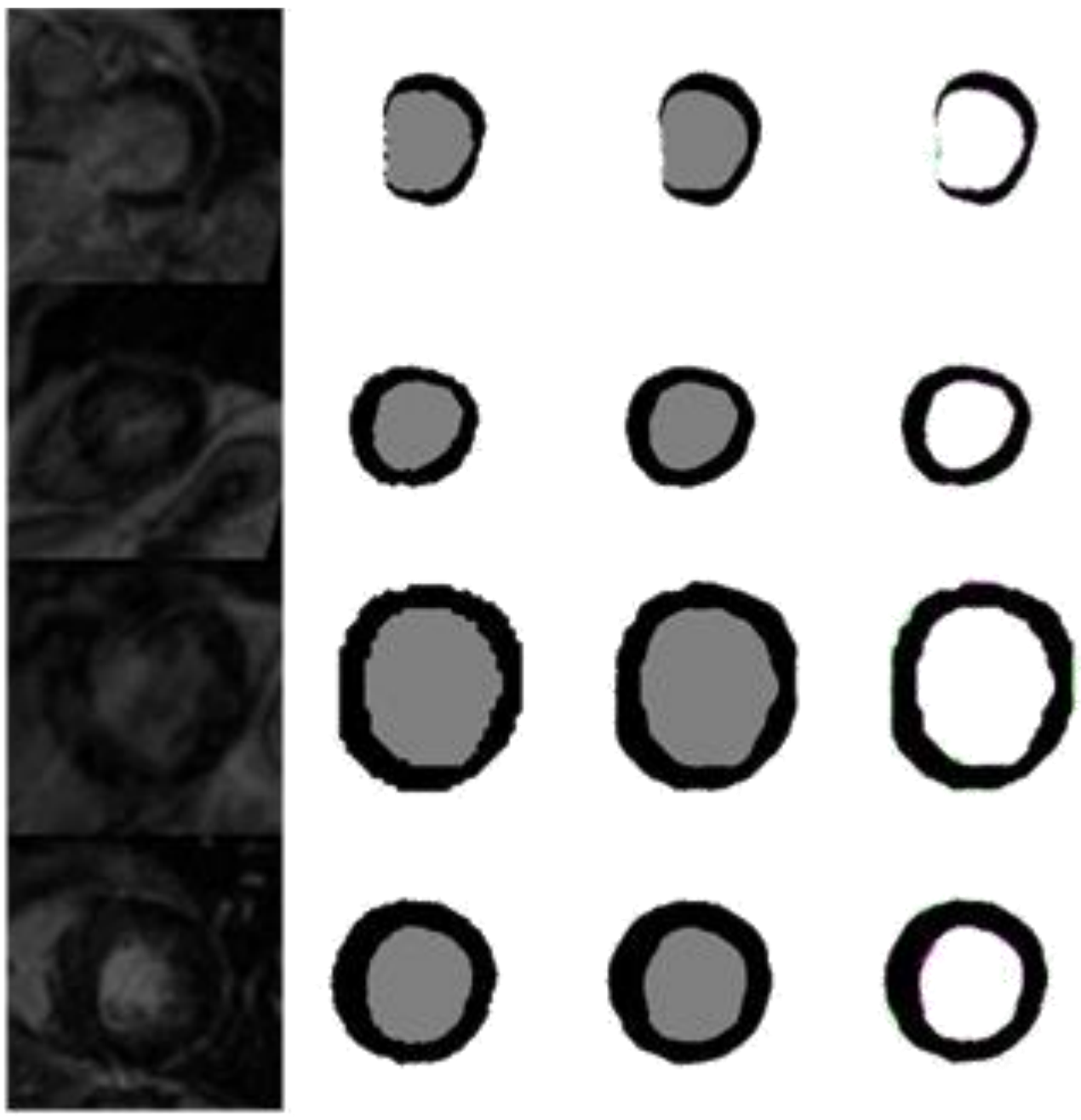

**Figure 16.** N256 segmented output masks | From top to bottom: Dice indices are equal to 0.9961, 0.9963, 0.9909, 0.9925.

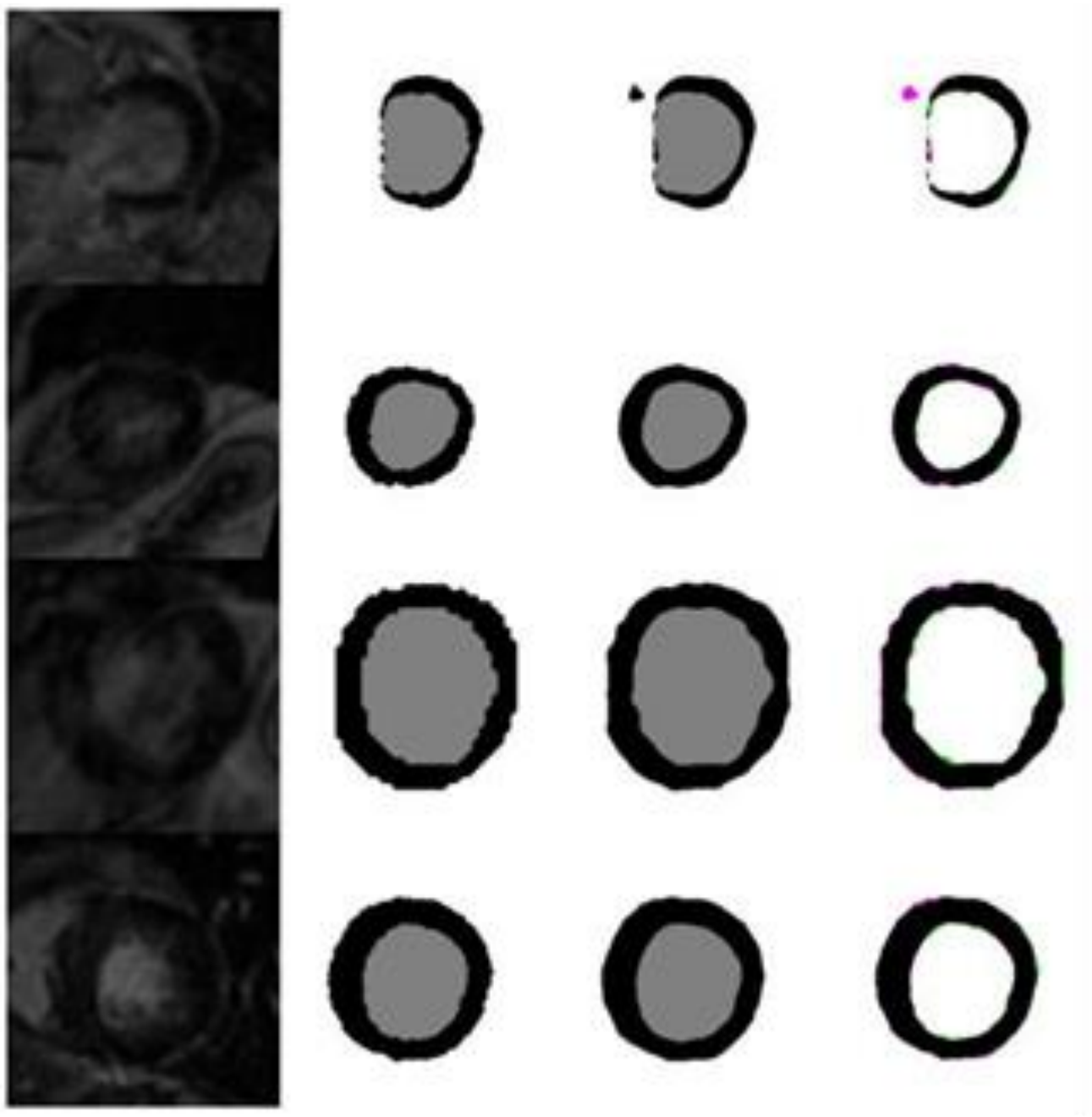

**Figure 17.** B256F segmented output masks | From top to bottom: Dice indices are equal to 0.9945, 0.9956, 0.9900, 0.9944.

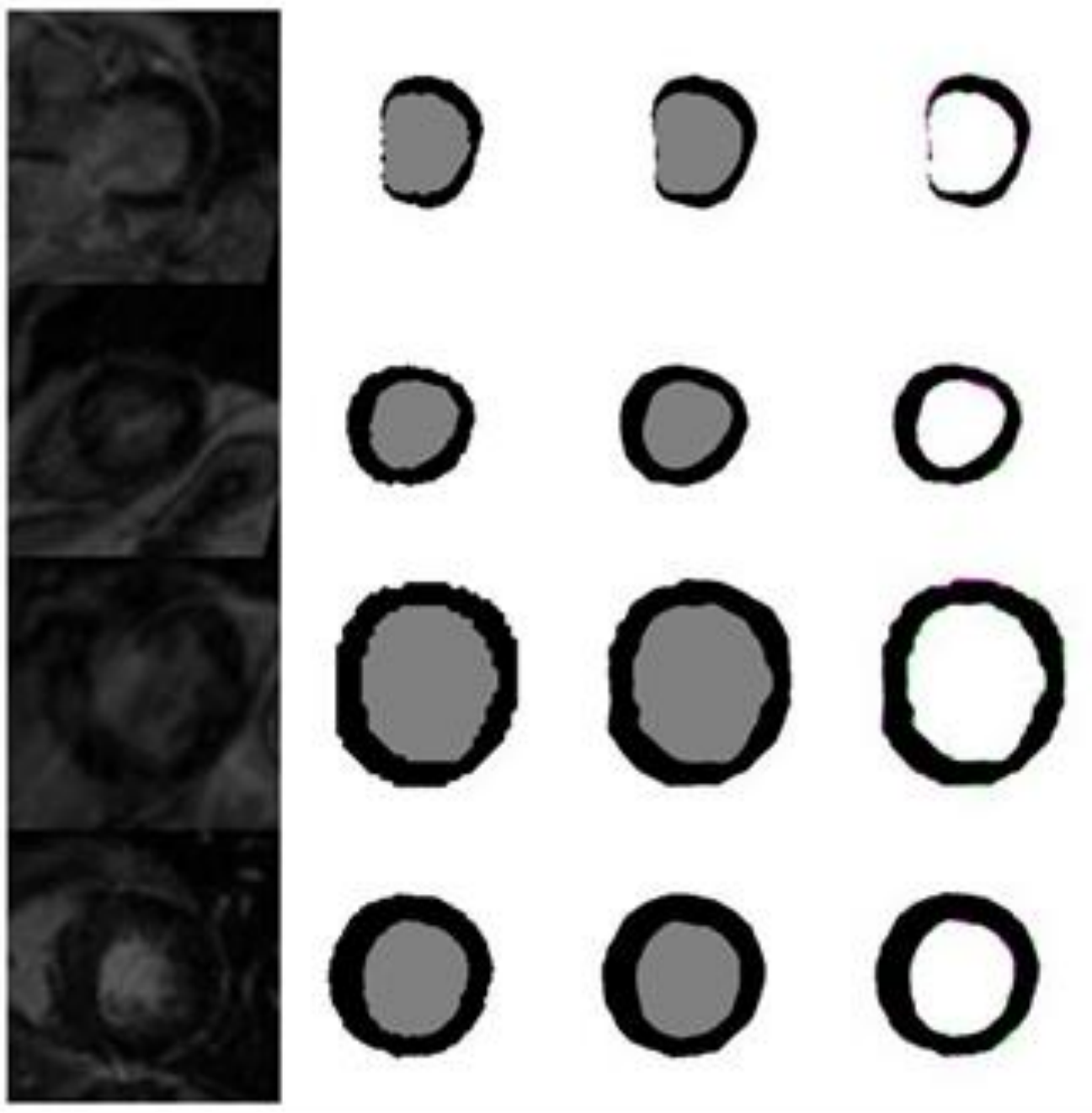

**Figure 18.** L256F segmented output masks | From top to bottom: Dice indices are equal to 0.9953, 0.9957, 0.9902, 0.9942.

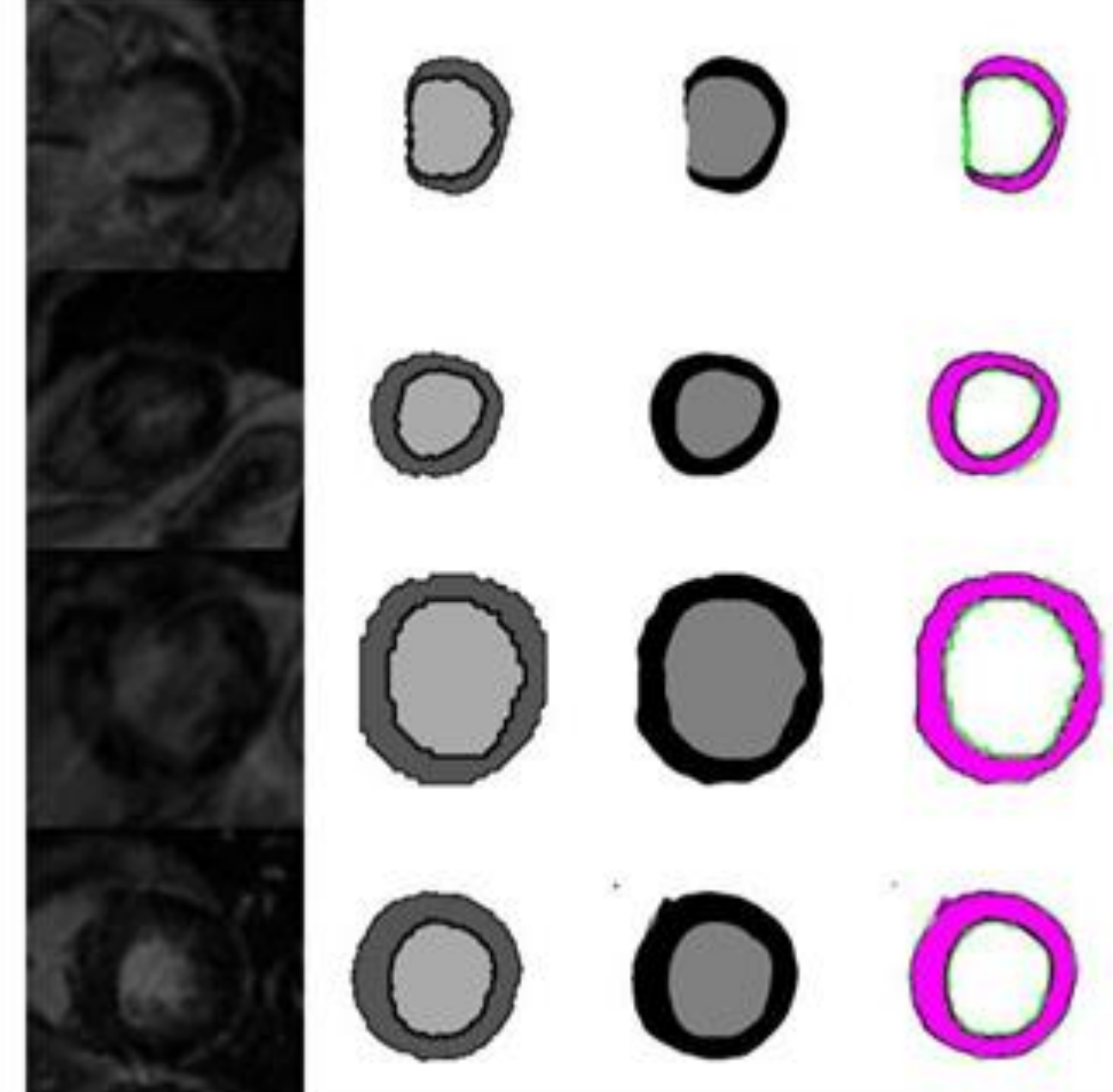

**Figure 19.** B256U segmented output masks | From top to bottom: Dice indices are equal to 0.9718, 0.9554, 0.8868, 0.9130.

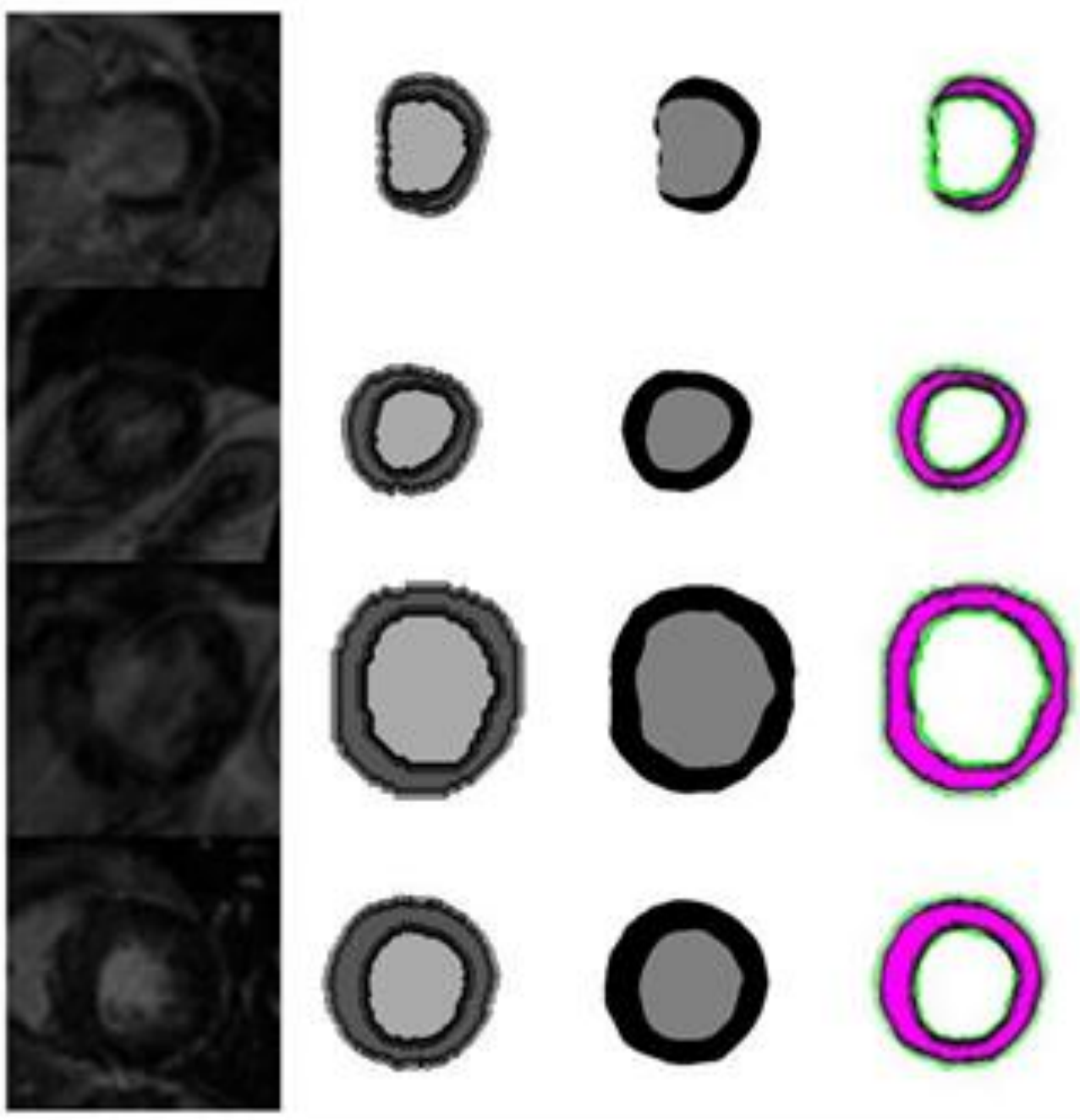

**Figure 20.** L256U segmented output masks | From top to bottom: Dice indices equal to 0.9694, 0.9558, 0.8854, 0.9150.

### *4.2. Evaluation of the effects of image size on the relationship between fully automated quantification and semi-automatic quantification of the MI results*

The arbitrary threshold, comparison of the sums, and sums of differences between medical experts or semi-automatic and fully automated quantification of MI results are three methods used to estimate the relationship, in terms of percentages, between semi-automatic and fully automated quantification of MI results. Here, it is important to note that the 100% percentage is the target percentage reflecting the semi-automatic or manual or medical expert-based results. Also, it is important to note that the MI quantification operation starts with an input image (resized to the size of interest), and fed through U-net which creates the corresponding output segmentation mask that is later analysed by EWA algorithm to produce MI quantification results.

#### 4.2.1. Arbitrary threshold

This method or strategy separates automated quantification of MI results using an arbitrary threshold or separate automated quantification results that are closer (to some extent) to manual or semi-automatic quantification results. With this option, threshold values, arbitrarily chosen, are 25, 15, and 0.35 for scar (ml), scar (%), and mo (%), respectively. These values reflect the author's opinion on the relationship strength or closeness between semi-automatic and fully automated quantification of the MI results. Here, it is important to note that other observers could have different opinions.

With this option, when the fully automated quantification results are less than 25, 15, and 0.35 for scar (ml), scar (%), and mo (%), respectively, the automated quantification results are close to some extent to manual or semi-automatic quantification results thus exists a strong or close relationship between semi-automatic and fully automatic quantification results. Table 3 shows the percentages, achieved using *option-1*, that help to estimate the relationship between semi-automatic or medical experts-based quantification (100%) and fully automated quantification (x %) results. In this context, the effects of image size on deep learning can be understood via how close achieved percentages are close to 100% in the cases of LGE-MRI images of the size 128 × 128 and 256 × 256, respectively.

**Table 3.** Percentages achieved using arbitrary threshold.

| Network | C128 | N256 | B256 | L256 |
|---|---|---|---|---|

| Scar (ml) | 87.5% | **91.6%** | **91.6%** | **91.6%** |
|---|---|---|---|---|
| Scar (%) | 79.1% | **95.8%** | 75% | 83.3% |
| MO (%) | 62.5% | 62.5% | 62.5% | **66.6%** |

4.2.2. Comparison of the sums

This method compares the sums of manual or semi-automatic and auto-mated results by calculating the percentage of the sum of scar (ml), scar (%), and mo (%) of manual results versus the percentage of the sum of scar (ml), scar (%) and mo (%) of fully automatic quantification results. Table 4 shows the percentages achieved, using *option 2*, that help to estimate, to some extent, the relationship between semi-automatic quantification (100%) and fully automated quantification (x %) results. Again, in this con-text, the effects of image size on deep learning can be understood via observing how close achieved percentages are close to 100% in the cases of LGE-MRI images of the size 128 × 128 and 256 × 256, respectively.

**Table 4.** Percentages achieved using comparison of the sums.

| Network | C128 | N256 | B256 | L256 |
|---|---|---|---|---|
| Scar (ml) | 58.4% | 49.5% | 72.2% | **72.3%** |
| Scar (%) | 74.8% | 75.1% | 74.7% | **75.7%** |
| MO (%) | 6.6% | 10.7% | **11.3%** | 9.5% |

4.2.3. Sums of differences

This method compares the sums of differences between semi-automatic and fully automated quantification of the MI results by calculating the percentage of the sum of differences of scar (ml), scar (%), and mo (%) of manual or semi-automatic results versus the percentage of the sum of differences of scar (ml), scar (%) and mo (%) of fully automatic results. Table 5 shows the percentages achieved, using option-3, that help to estimate, to some extent, the relationship between medical experts-based or semi-automatic quantification of MI (100%) and fully automated quantification of MI (x %) results. Like in the previous two options cases, effects of image size on deep learning are also demonstrated by such percentages and can be understood via observing how close achieved percentages are close to 100% in the cases of LGE-MRI images of the size 128 × 128 and 256 × 256, respectively.

**Table 5.** Percentages achieved using sums of differences.

| | C128 | N256 | B256 | L256 |
|---|---|---|---|---|
| Scar (ml) | **78.2%** | 74.4% | 72.4% | 74.8% |
| Scar (%) | **79.07%** | 75.2% | 71.9% | 73.7% |
| MO (%) | 74.2% | 75.2% | 75.08% | **75.4%** |

To better interpret the results, presented in Table 3, Table 4, and Table 5, it is important to bring attention to the following: In each of the three tables, each U-net has a maximum of 3 chances of outperforming the rest in terms of scar (ml), scar (%), and mo (%). In three tables, the total chances for each U-net increase to 9 times per each U-net. As can be seen, via bolded percentages, in Table 3, Table 4 and Table 5, C128, N256 and B256 achieved the highest percentage 2 times over 9 expected – which is equivalent to 22.2%. However, L256 achieved the highest percentage 5 times over 9 expected – which is equivalent to 55.5%. With this in mind – quantification results (i.e., the highest or best) based on the dataset of bigger LGE MRI images are 55.5% closer the manual or semi-automatic results while quantification results based on the dataset of smaller LGE MRI images are 22.2% closer the manual results. It is important to note that the Segment CMR software's EWA algorithm is responsible for generating the scar (ml), scar (%), and mo (%) values, (including possible quantification errors) once the plugin of interest or plugin linked to the trained U-net is run. Therefore, it is important to note that possible annotation and EWA algorithm errors may significantly affect results in this context – meaning that future works must pay attention to the effects of those possible sources of fully automatic quantification errors.

4.2.4. Comparison of the results from semi-automatic and fully automated quantification of MI

As can be seen in Figure 21, Figure 22, and Figure 23, twenty-four stacks of LGE-MRI images, referred to as CHIL-2-6-xxxxx, were used during the experiments. Also, these figures graphically show the variation of results from two main quantification approaches, namely: *semi-automatic* (manual) and *fully automated* (C128, N256, B256, L256).

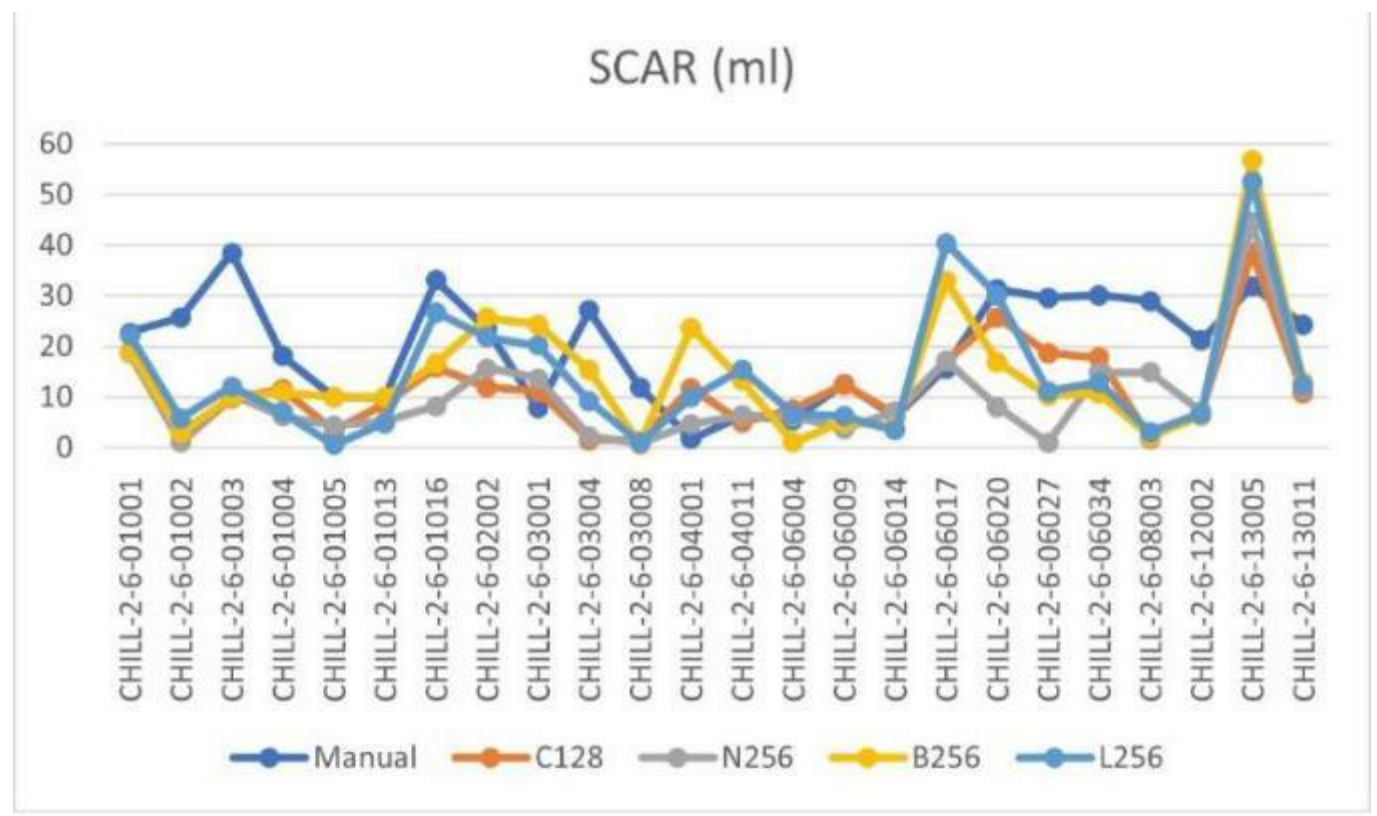

**Figure 21.** MI quantification results - scar (ml).

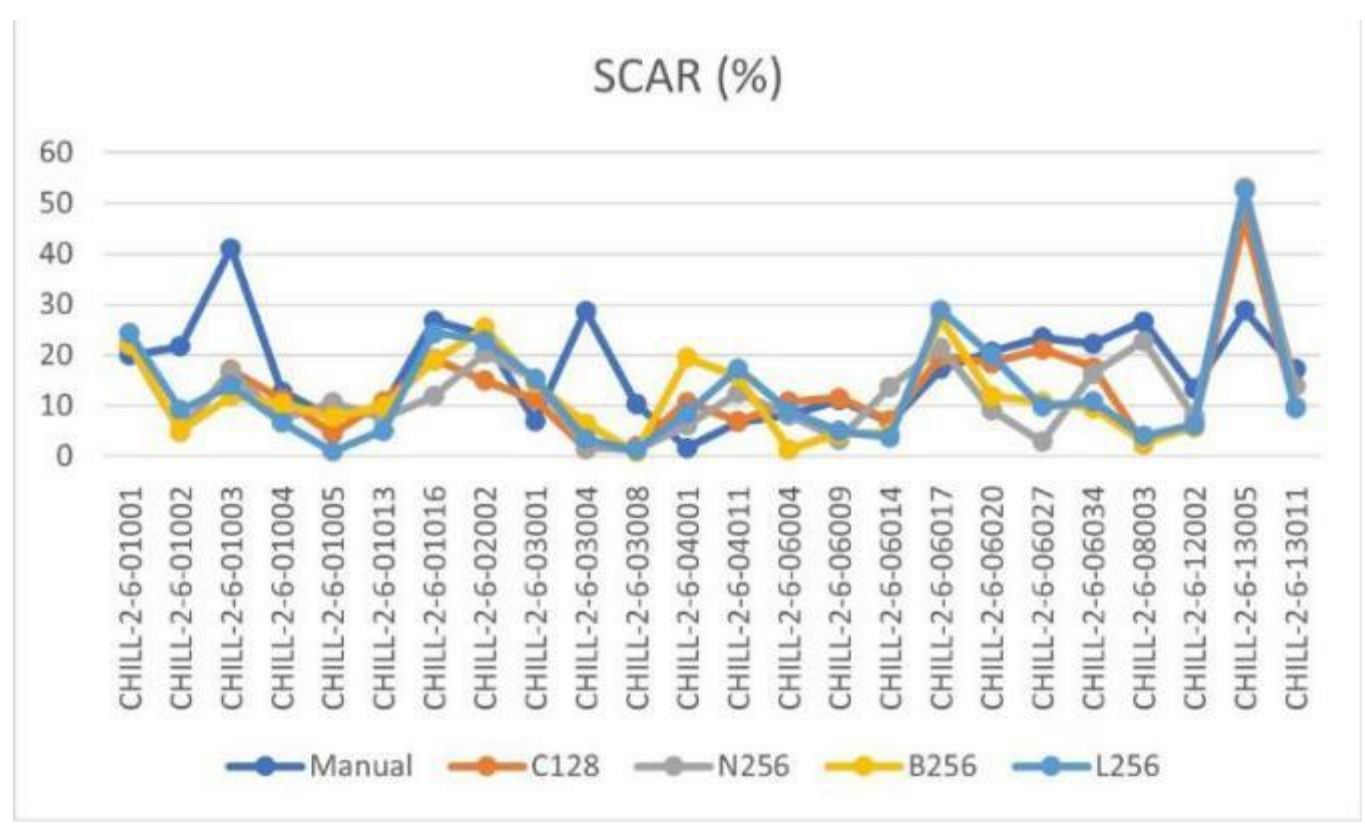

**Figure 22.** MI quantification results - scar (%).

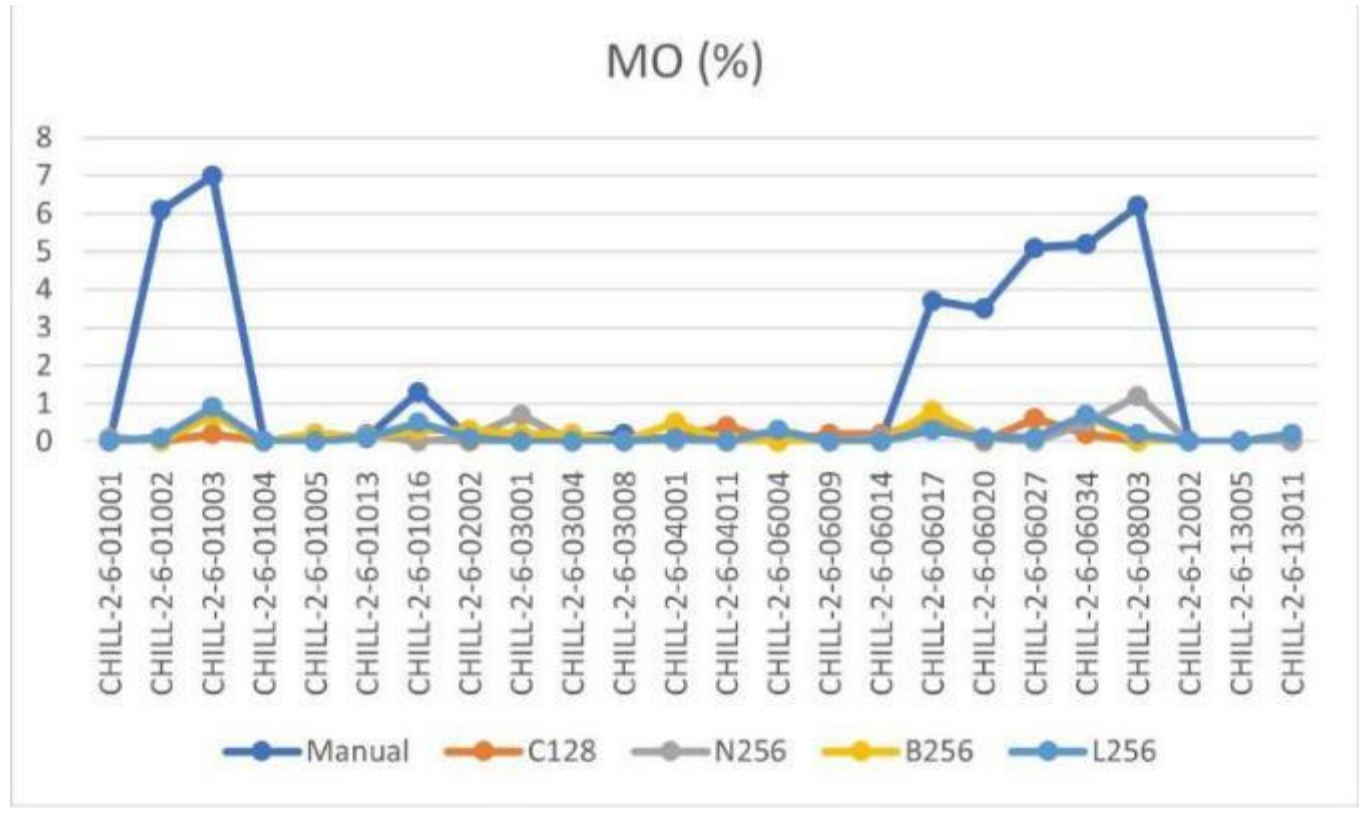

**Figure 23.** MI quantification results - mo (%).

## 5. Conclusions

Effects of the size of LGE-MRI images for training datasets were investigated, presented and discussed. Specifically, such effects were presented in terms of the quality of automatic seg-mentation with U-net against the GT segmentation and the relationship between fully automated quantification and semi-automatic quantification of MI results. After conducting experiments, a close relationship between semi-automatic and fully automated quantification of MI results was more detected in the case involving the dataset of bigger LGE MRI images than in that of the dataset of smaller LGE-MRI images. This happened be-cause the outputs of the U-net trained on LGE-MRI images of the size 256 × 256 were much closer to target vectors than the U-net trained on LGE-MRI images of the size 128 × 128. In other words, the cross-entropy loss in U-net trained on the training set of LGE-MRI images of the size 256 × 256 was lower than in U-net trained on the training set of LGE-MRI images of the size 128 × 128 - while it was well known that the lower the loss, the more accurate the model (i.e., U-net in this case). U-nets trained on the training set of LGE-MRI images of the size 256 × 256 took more time than U-net trained on the training set of LGE-MRI images of the size 128 × 128.

It is important to note that, the study main objective was to determine the best size for LGE-MRI images in the training dataset that could contribute to the improvement of LGE MRI image segmentation accuracy. Also, seeking to determine the best size and improve the segmentation accuracy required the use of extra-pixel category-based image interpolation algorithms instead of the traditional nearest neighbor of the non-extra pixel category. Given that extra pixel category interpolation algorithms produced extra class labels in the GT masks, this problem required the development of a novel strategy to remove extra class labels in interpolated GT segmentation masks. Finally, experimental results were provided to show how the change-in-LGE-MRI-image-size improved or worsened predictive capability or performance of U-net via segmentation and subsequent MI quantification operations. Note that, prior experiments the author conducted demonstrated that interpolating training samples or images using an extra-pixel category-based interpolation algorithm and interpolating masks using the nearest neighbor interpolation algorithm did not produce the results superior to cases of experiments shown in this paper where the same interpolation algorithm was used for images and masks. Note that, this study introduced a new way for interpolation mask handling or processing. Further research is needed to address potential errors in training datasets annotations and investigate errors in the EWA algorithm.

**Funding:** This research work was supported by Lund University between July and December 2020

**Data Availability Statement:** Data supporting the conclusions of this paper are not made public but are available on request and approval.

**Acknowledgments:** The author would like to thank Lund University and Medviso for the materials. Also, the author would like to thank reviewers and editors for their helpful comments.

**Conflicts of Interest:** The author declares no conflict of interest. The funders had no role in the design of the study; in the collection, analyses, or interpretation of data; in the writing of the manuscript; or in the decision to publish the results.